\documentclass[a4paper,fleqn]{cas-dc}

\usepackage{tabularx}
\usepackage{dblfloatfix,caption,afterpage}
\usepackage{multirow, makecell}
\usepackage{multicol}
\usepackage{array}
\usepackage{mathtools}
\usepackage{xurl}
\usepackage{nicefrac, xfrac}
\usepackage{enumitem}
\usepackage{subcaption}
\usepackage{placeins}
\usepackage{float}
\usepackage{wrapfig}

\usepackage[authoryear]{natbib}

\def\tsc#1{\csdef{#1}{\textsc{\lowercase{#1}}\xspace}}
\tsc{WGM}
\tsc{QE}

\begin{document}
\let\WriteBookmarks\relax
\def\floatpagepagefraction{1}
\def\textpagefraction{.001}

\shorttitle{Supervised learning of spatial features with STDP and homeostasis}    

\shortauthors{Davies, S. et al.}  

\title [mode = title]{Supervised learning of spatial features with STDP and homeostasis using Spiking Neural Networks on SpiNNaker}

\tnotemark[1] 

\tnotetext[1]{Using STDP and homeostasis on spiking neural networks simulated on SpiNNaker, Davies et al. demonstrate that it is possible for such a network to learn and recognise spike patterns by presenting the desired pattern to the network only once. The pattern is presented as a set of simultaneous spikes at the input layer, and the output is produced after a short delay. In addition, the same network is trained with multiple patterns, and the accuracy and other performance metrics are computed.} 

%

\author[1]{Sergio Davies}[type=author, prefix=Dr, orcid=0000-0001-5330-5527]

\cormark[1]

\fnmark[1]

\ead{sergio.davies@mmu.ac.uk}

\ead[url]{https://www.mmu.ac.uk/staff/profile/dr-sergio-davies}

\credit{Conceptualisation of this study, Design of the methodology, Software development, Interpretation of the results, Preparation of the article}

\cortext[1]{Corresponding author}

\affiliation[1]{organization={Department of Computing and Mathematics, Manchester Metropolitan University},
            addressline={John Dalton Building, All Saints Campus, Oxford Road}, 
            city={Manchester},
            postcode={M15 6BH}, 
            state={United Kingdom},
            country={U.K.}}



\author[2]{Andrew Gait}[type=author, prefix=Dr]




\credit{Support on the definition of the methodology and on the interpretation of the results, Software development, Preparation of the article}

\author[2]{Andrew Rowley}[type=collab, prefix=Dr]




\credit{Software development, Review of the article}

\affiliation[2]{organization={APT group, School of Computer Science, The University of Manchester},
            addressline={IT Building, Oxford Road}, 
            city={Manchester},
            postcode={M13 9PL}, 
            state={United Kingdom},
            country={U.K.}}

\author[3]{Alessandro Di\ Nuovo}[type=collab, prefix=Prof.]




\credit{Review of the article}

\affiliation[3]{organization={Department of Computing, Sheffield Hallam University},
            addressline={Cantor Building, 153 Arundel Street}, 
            city={Sheffield},
            postcode={S1 2NU}, 
            state={United Kingdom},
            country={U.K.}}

\begin{abstract}
Artificial Neural Networks (ANN) have gained significant popularity thanks to their ability to learn using the well-known backpropagation algorithm.
Conversely, Spiking Neural Networks (SNNs), despite having broader capabilities than ANNs, have always posed challenges in the training phase.
This paper shows a new method to perform supervised learning on SNNs, using Spike Timing Dependent Plasticity (STDP) and homeostasis, aiming at training the network to identify spatial patterns. Spatial patterns refer to spike patterns without a time component, where all spike events occur simultaneously.
The method is tested using the SpiNNaker digital architecture.
A SNN is trained to recognise one or multiple patterns and performance metrics are extracted to measure the performance of the network.
Some considerations are drawn from the results showing that, in the case of a single trained pattern, the network behaves as the ideal detector, with $100\%$ accuracy in detecting the trained pattern.
However, as the number of trained patterns on a single network increases, the accuracy of identification is linked to the similarities between these patterns.
This method of training an SNN to detect spatial patterns may be applied to pattern recognition in static images or traffic analysis in computer networks, where each network packet represents a spatial pattern.
It will be stipulated that the homeostatic factor may enable the network to detect patterns with some degree of similarity, rather than only perfectly matching patterns.
The principles outlined in this article serve as the fundamental building blocks for more complex systems that utilise both spatial and temporal patterns by converting specific features of input signals into spikes. One example of such a system is a computer network packet classifier, tasked with real-time identification of packet streams based on features within the packet content.
\end{abstract}




\maketitle

\section{Introduction}
The rising popularity of neural networks can be attributed to their information processing capabilities, despite being regarded as ``black-box'' systems due to their emulation of the behaviour of biological neural networks, rather than relying on established biological structures \citep{Prieto2016NeuralChallenges}.

Artificial neural network models draw inspiration from their biological counterparts, attempting to mimic how the human brain performs specific tasks \citep{Haykin1999NeuralFoundation}.
Based on the computational units (neurons) used in these networks, we can classify three main categories of neural networks \citep{Maass1997NetworksModels}.
Each of these categories is referred to as a ``generation'' of neural networks. Each generation simulates biological processes with an increasing degree of accuracy.

The first generation of neural networks were dominated by the McCulloch-Pitts neuron model \citep{McCulloch1943AActivity} which allows discrete inputs and outputs (only ``0''s or ``1''s).
The next generation (second generation of neural networks, more commonly known as Artificial Neural Networks) evolved this model to allow input and output values to be continuous within a specified range, either $[0;1]$ or $[-1;1]$.

In both these generations of neural networks, the output of a neuron is transferred to the subsequent neuron(s) through weighted connections.
This weight is altered during the training phase by presenting the network with an input for the training session, analysing the network output, and determining the error between the current network output and the desired output.
This error is then used to compute the changes in synaptic weights throughout the network using the backpropagation algorithm \citep{Linnainmaa1976TaylorError, Schmidhuber2015DeepOverview}.

It is suggested that the popularity of Artificial Neural Networks (ANNs) can be attributed to the use of the backpropagation training algorithm \citep{Lecun2015DeepLearning}.
This algorithm has played a pivotal role in enabling the training of significant ANN models \citep[e.g.:][]{Sermanet2013PedestrianLearning,Mohamed2012AcousticNetworks,Dahl2012Context-dependentRecognition}.
Indeed, backpropagation has provided an advantage to this type of networks with a well-known and robust method of training which has now been embedded in most, if not all, ANN simulation platforms.

However, both the first and the second generations of neural networks do not consider one fundamental aspect: biological networks evolve following biological time.
Artificial neural networks perform their operations in abstract time that does not correspond to biological time.
Even advanced models such as Continuous Timescale Recurrent Neural Networks \citep[CTRNN,~][]{Funahashi1993ApproximationNetworks} and Multi-Timescale Recurrent Neural Networks \citep[MTRNN,~][]{Yamashita2008EmergenceExperiment} use the recurrent structure of the neural network to keep track of the state and its evolution. However, this does not correspond to biological real-time, but rather follows the number of iterations in the network.

The third generation of neural networks \citep{Maass1997NetworksModels}, also known as Spiking Neural Networks (SNNs), improves the biological realism of previous generations of neural and synaptic models by introducing the time variable in the models \citep{Gerstner2014NeuronalDynamics, Kasabov2019}.
Indeed, the models proposed for this generation of neural networks are directly inspired from biology: the most realistic model is the Hodgkin-Huxley neuron \citep{Hodgkin1952ANerve}, which is also the most complex to simulate numerically on a computer.
Other models, instead, limit their biological plausibility to reduce their numerical complexity \citep{Izhikevich2004WhichNeurons}.
All of these neuron models are described using differential equations that depict the evolution of a neuron's state over time \citep[e.g.:][]{Izhikevich2003SimpleNeurons, Lapicque1907RecherchesPolarization, Dayan2001TheoreticalSystems, Kasabov2019}.

The communication between neurons is also inspired from biology: it is known that neurons interact by means of action potentials, also known as spikes, \citep{Dayan2001TheoreticalSystems}.
Such communications use a wide array of mechanisms to encode information, as described by \cite{Auge2021ANetworks}, and even more methods could be envisioned.

Similarly, also synapses (the interconnection between neurons) in second generation neural networks are represented by a single number that represents its ``strength''.
This value is altered during training using the backpropagation algorithm.
However, from a biological perspective, this algorithm has raised some skepticism on its plausibility \citep{Tavanaei2019DeepNetworks}.
This is also supported by the fact that biological findings have shown that signals transmitted through synapses have a time evolution that may be described through a differential equation \citep{Kasabov2019}.

In addition, biology has described a number of mechanisms that allow biological neural networks to adapt to input stimuli.
Among these we can mention STP -- short-term plasticity \citep{Zucker2002Short-TermPlasticity}, STDP -- Spike Timing Dependent Plasticity \citep{Markram1997RegulationEPSPs,Bi2001SynapticRevisited}, homeostasis \citep{Turrigiano1999HomeostaticSame,Turrigiano2011TooRefinement}, structural plasticity \citep{Zecevic1991SynaptogenesisCortex} and evolutionary learning \citep{Schmidhuber2015DeepOverview}.
All these mechanisms, following different processes, alter the architecture of the neural network by altering the synaptic strength of existing synapses, by creating new synapses (synaptogenesis), or by removing existing synapses (synaptic pruning).
Despite all the changes imposed by these mechanisms, neurons within the network need to maintain their functional stability, and the network itself needs to keep a stable behaviour.
This happens through the homeostatic process at two levels: on a neuron scale it helps to keep a healthy neural activity, while on the network scale homeostasis helps keeping the network stability \citep{Maffei2009NetworkCoordination}.

The learning process underlying mechanism was proposed by \cite{Hebb1949TheTheory}, and commonly summarised as \emph{``Cells that fire together wire together''}.
More details of this biological process have emerged in the last few decades, leading to a number of learning rules which can affect synapses on a time interval spanning a few milliseconds to a lifetime, or more, through generations of individuals \citep{Zeraati2021Self-OrganizationPlasticity, Nolfi1994LearningNetworks}.

As a general rule, the longer the learning period, the more permanent the effects are on the neural networks: short-term plasticity affects quickly the stability of the network, but the effects do not last very long \citep{Tsodyks2013Short-termPlasticity}.
On the other hand, long-term plasticity has a stronger impact on the network, so that it allows the network to self-organise towards a stable critical regime \citep{Zeraati2021Self-OrganizationPlasticity}.
Evolutionary learning has an even stronger impact that allows generation of individuals to behave in a specific way innately \citep{Tierney1986TheEvolution}.

Such learning rules have been replicated in computer simulations, and showed their characteristics in applied tasks.
In particular, it is relevant to mention that STDP was successfully applied in many applications related to the identification of spatio-temporal spike patterns \citep[e.g.:][]{Davies2012LearningNetworks, Guyonneau2005NeuronsSTDP, Masquelier2008SpikeTrains, Davies2012AImplementation}.

A neural network is trained within an environment.
On the basis of this it is possible to classify four learning paradigms \citep{Kasabov2019} depending on the presence and the structure of the teaching signal: supervised learning, semi-supervised learning, unsupervised learning and reinforcement learning.
The learning rules introduced before (STP, STDP, etc.) refer to an unsupervised learning paradigm, where the teaching signal is absent and the network aims to identify autonomously a pattern in the input signal.

In this paper we present a novel method of training a spiking neural network to identify spatial patterns (patterns of spikes presented at the same time as input to the network) using STDP and homeostasis: two learning algorithms acting on different time-scales collaborating to achieve a task.
The network is trained initially to identify a single pattern and the accuracy is then evaluated by testing exhaustively all the possible input patterns to the network.
In a second step, the network is trained to identify two patterns, and we will show that the accuracy of the identification depends strictly on the degree of similarity between the two patterns on which the network has been trained. This similarity will be measured by the Hamming Distance between input patterns.
Finally, a more thorough experiment includes training the network on three patterns and measuring again the detection accuracy, among other classification metrics.

The experiments are performed on the SpiNNaker digital architecture \citep{Furber2020}, using the sPyNNaker implementation of the PyNN neural network language \citep{Rhodes2018SPyNNaker:SpiNNaker,Rowley2019SpiNNTools:Platform}.
SpiNNaker is a system designed at the University of Manchester: each SpiNNaker chip comprises eighteen very-low-power ARM986 processors (cores); the main SpiNNaker server, housed at the University of Manchester, consists of a million cores built of multiple boards containing multiple chips.
Computations in the brain are inherently parallel and the architecture is designed to mimic this parallelism.
SNNs may be simulated on the machine by submitting scripts based on the PyNN neural network language. These scripts are then converted by the software stack into executable files which run on as many cores as required by the neural network.

The remaining sections of this article encompass a detailed account of the experiments outlined in the methodology section (Section \ref{label:methodology}).
This includes an in-depth exploration of the STDP learning rule (Section \ref{label:STDP}), training procedures (Section \ref{label:TrainingPhase}), and testing methods (Section \ref{label:TestingPhase}).
Subsequently, the results section (Section \ref{label:results}) will shed light on the research outcomes, involving training the network on a single pattern (Section \ref{label:single_pattern}), two patterns (Section \ref{label:dual_pattern}), and multiple patterns (Section \ref{label:multi_pattern}).
In each case, various classification metrics will be employed to assess the network's performance in pattern identification following the training process.
Finally, the conclusion section (Section \ref{label:conclusions}) will summarise the key findings of this research and its applicability.

While the architecture of the neural network proposed in this paper may appear simplistic, it is intentionally designed as such to study the training process outlined within this research and isolate each component's effects on the network's performance.

\section{Methodology}\label{label:methodology}

In this paper we refer to spatial patterns of spikes as a set of spikes that are presented to the network from different source neurons at the same time, and whose source neuron is meaningful for the pattern.

As spatial patterns relate only to the presence or absence of a spike from a specific source, this type of patterns can be identified and encoded with the use of binary numbers, where ``1''s represent the presence of a spike, while ``0''s reflect its absence.
These numbers represented either in their binary or decimal format will also be referred to as ``code words''.

The spikes used to transfer information follow two of the possible encodings suggested by \cite{Auge2021ANetworks}, namely:

\begin{itemize}
    \item \textbf{Time To First Spike}: This is utilised during the network training phase to ensure that the supervised learning paradigm correctly triggers the relevant side of the STDP learning rule. Depending on whether the input spike represents  ``0'' or ``1'', the generated spike occurs slightly before or after the training signal.
    \item \textbf{Parallel binary encoding}: Since spatial patterns only correspond to the presence or absence of spikes from each source, we can represent patterns with binary numbers that are presented to the network. A binary number ``1'' indicates the presence of a spike from that source, while ``0'' represents its absence.
\end{itemize}

Two related neural networks are designed for this exercise: the first one is used to train the relevant synapses, while the second network, which is a simplified version of the training network, is used to test and validate the model obtained in the first step. 

\subsection{STDP on SpiNNaker}\label{label:STDP}

\begin{figure}[tp]
\centering
\includegraphics[width=\linewidth]{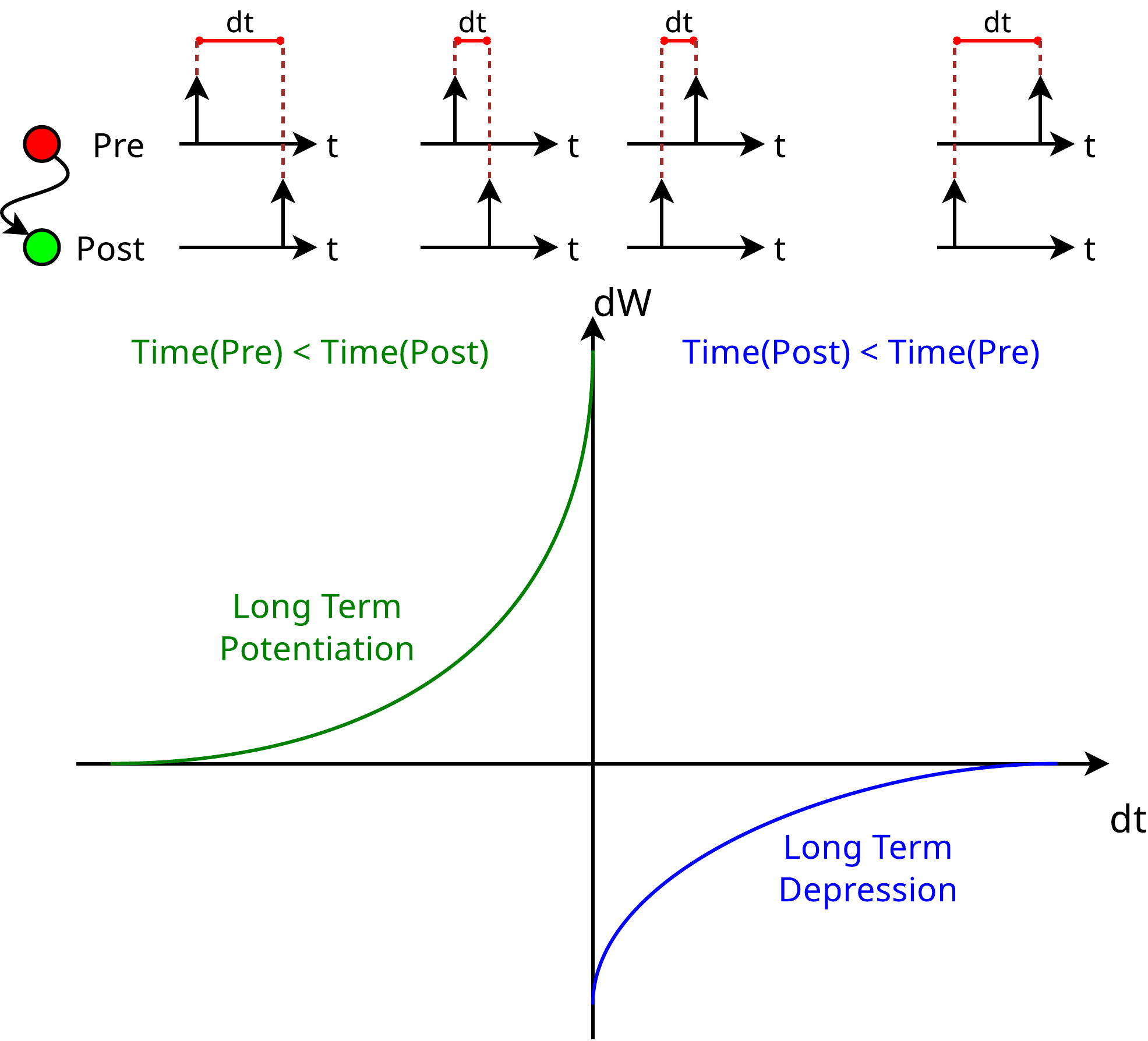}
\caption{An example of how weight change (dW) is calculated based on the time difference between pre- and post-synaptic spikes (dt). In green on the left is the Long Term Potentiation (LTP) generated by a pre-synaptic and post-synaptic spike sequence. In blue on the right is the Long Term Depression (LTD) generated by a post-synaptic and pre-synaptic spike sequence.}
\label{fig:STDP}
\end{figure}

\begin{figure*}[!tp]
\centering
\includegraphics[width=\linewidth]{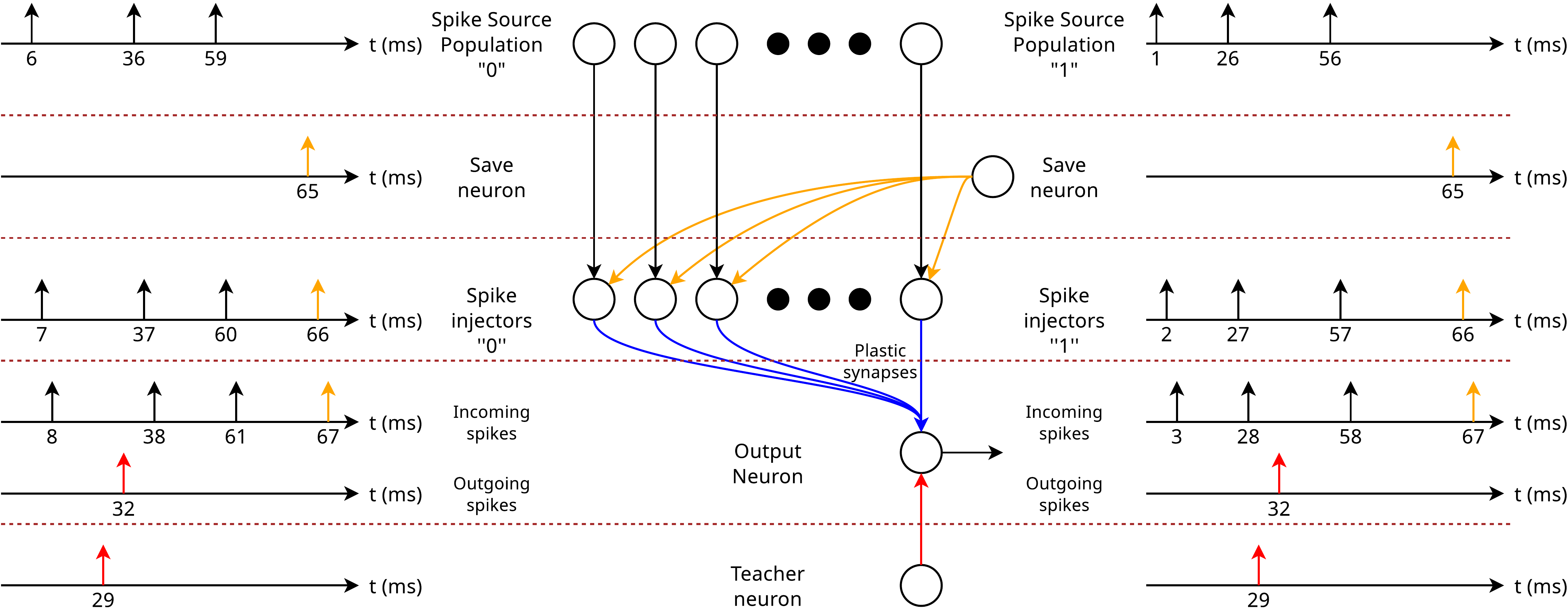}
\caption{The sequence of spikes in the network used for training.}
\label{fig:trainingTimesFull}
\end{figure*}

\begin{figure}[!bp]
\centering
\includegraphics[width=\linewidth]{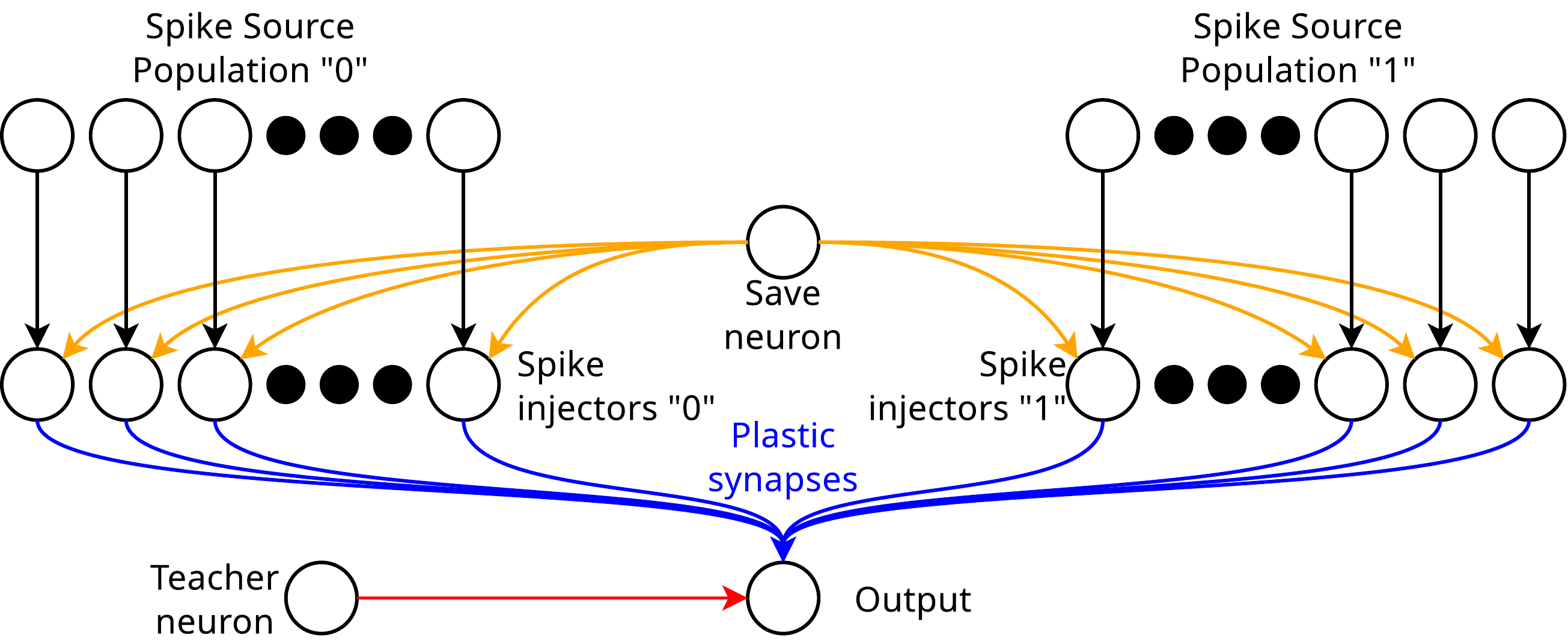}
\caption{The network used for training. In blue the STDP-enabled synapses.}
\label{fig:TrainingNetwork}
\end{figure}

STDP is a form of learning whereby the weight of a synapse between two neurons is either potentiated (LTP) or depressed (LTD) dependent upon whether a post-synaptic spike follows or precedes a pre-synaptic spike.
The size of this change, in general, drops off exponentially as the time difference between the pre- and post-synaptic spikes gets larger \citep{Sjostrom2010Spike-timingPlasticity}. This is shown graphically in fig.\ref{fig:STDP}.
In PyNN, STDP is defined in a modular fashion such that the user may specify which timing rule (for example, to determine the shape of the exponential decay) and weight update rule (for example, to indicate whether the weight update is additive or multiplicative) they wish to use.

This is how the rules are also implemented on SpiNNaker, with one proviso: due to local memory restrictions on how much data can be held for parameters, multiple STDP projections to the same target population must use the same rule with the same parameters.

On SpiNNaker, the plasticity mechanism for STDP is also only activated when the post-synaptic neuron receives the second (pre-synaptic) spike: at least two pre-synaptic spikes are, therefore, required for the calculations to take place.
This is because the conventional method for calculating STDP at every pre-synaptic spike and every post-synaptic spike is difficult on SpiNNaker due to the synaptic weights being held in external memory and only copied into local memory when a pre-synaptic spike arrives.
Thus, a deferred event-driven model is used to postpone the STDP calculation until future spike timings determine how the pre-synaptic sensitive scheme is applied \citep{Jin2010ImplementingHardware}.
Because of this deferred event-driven model, STDP weight changes can only be computed when a pre-synaptic spike is received.
Therefore, to detect the effects of the sequence of spikes to the output neuron at the end of the training phase, a ``save neuron'' (see fig.\ref{fig:trainingTimesFull} and fig.\ref{fig:TrainingNetwork}) emits a final spike whose only effect is to trigger the execution of the STDP learning rule on plastic synapses.

\subsection{Training phase}\label{label:TrainingPhase}

The training phase relies on the network shown in fig.\ref{fig:TrainingNetwork}.
This network consists of two sections: one focused on the ``0''s on the left, and, symmetrically, another section dedicated to the ``1''s on the right.

The ``Spike Source Populations'' inject spikes according to specific patterns to train the network. The ``Spike injector'' populations comprise leaky integrate-and-fire neurons with delta synapses.
These synapses have the characteristic that the current transferred to the post-synaptic neuron is applied within a single-millisecond time slot, during which it receives all the current.
The neuron parameters are set to the default values provided by the PyNN \citep{Davison2009PyNN:Simulators} interface to the SpiNNaker backend simulator.

Fig.\ref{fig:trainingTimesFull} shows the spike times for each neuron in the network in the case of a ``0'' on the left and in case of a ``1'' on the right.
The information is encoded using the ``Time to first spike'' method: in case a ``0'' needs to be presented, the sequence of spikes generated by the Spike Source Population ``0'' includes spikes at $6$, $36$ and $59$ milliseconds, while to encode a ``1'' the Spike Source Population ``1'' emits spikes at $1$, $26$ and $56$ milliseconds. These spikes are propagated through to the output neuron following the time pattern in fig.\ref{fig:trainingTimesFull}. Indeed, the neurons in the ``Spike Injector'' populations emit a spike for each spike they receive.

The output neuron receives the features of the signal from the plastic synapses (in blue) which do not contribute to the membrane potential since their weight is $0$. However, they allow the output neuron to store information to trigger the STDP learning rule. This sequence of spikes has been designed considering the peculiarities of both the SpiNNaker architecture and of the software implementation of the STDP algorithm \cite{Jin2010ImplementingHardware}. The STDP algorithm employed in this case is the nearest neighbour spike pair rule, which is only triggered when at least one pre-synaptic spike and one post-synaptic spike are already in memory at the point when an new incoming pre-synaptic spike is received. This is a custom extension for the SpiNNaker backend simulator in PyNN. The parameters used for the weight update rule are as follows: $\tau_+=5$, $\tau_-=5$, $A_+=1$ and $A_-=-1$. However, later in this article it will be discussed that the specific values of these parameters have little relevance on the whole set of experiments.

\begin{figure}[tp]
\centering
\includegraphics[width=\linewidth]{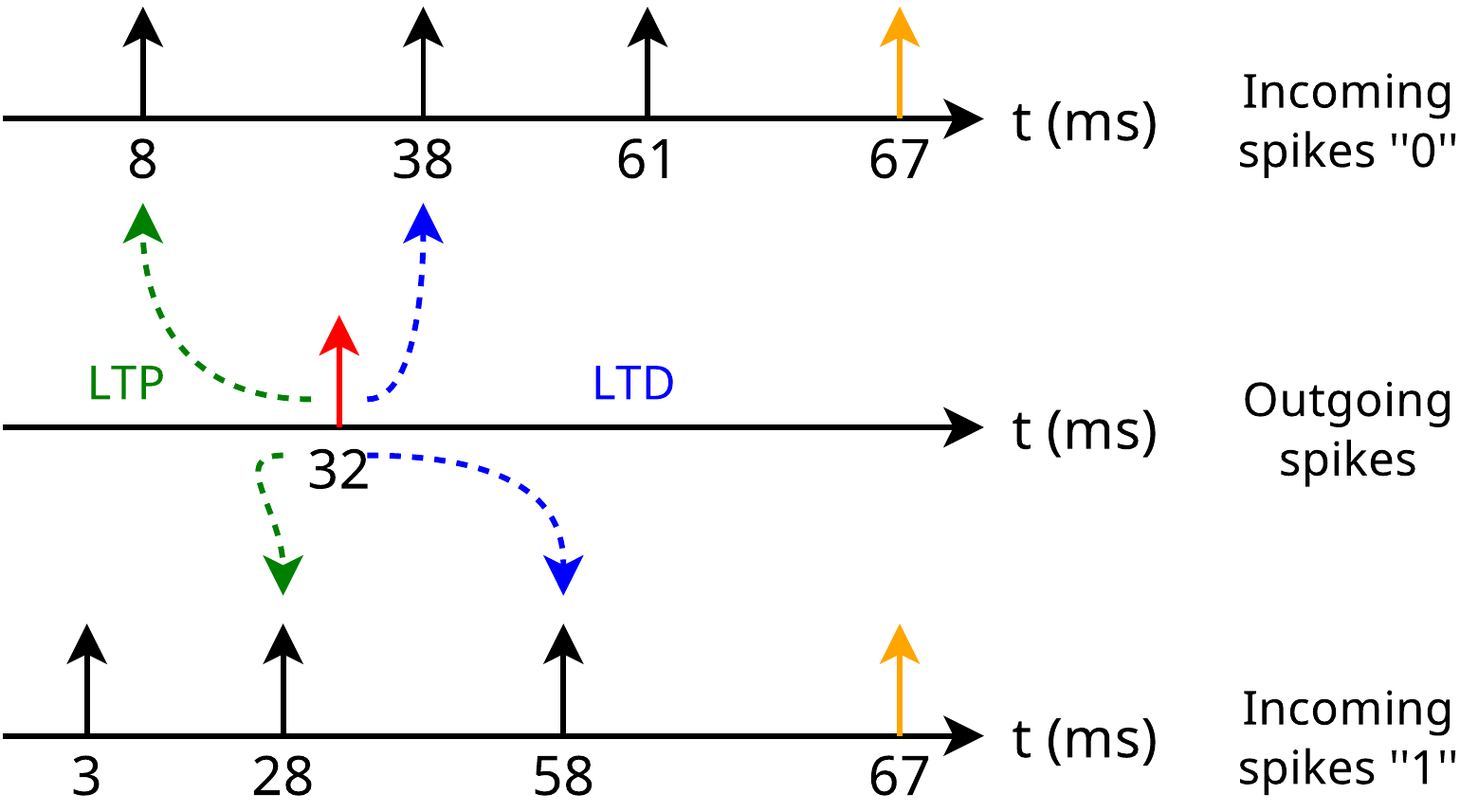}
\caption{The precise timing of the spikes for potentiation and depression. The timing for both Long Term Potentiation (LTP) and Long Term Depression (LTD) is considered on the timestep immediately following the outgoing spike value, so the values concerned here are 5 ms when the "1" is potentiated and the "0" is depressed, and 25 ms when the "0" is potentiated and the "1" is depressed.}
\label{fig:preciseTiming}
\end{figure}

In the model presented above, the output spike generated by the signal from the teacher neuron (in red) is always received at millisecond $31$ and generates an output spike at millisecond $32$. This output spike always falls between at least two pre-synaptic spikes, both in the case of a ``0'' and a ``1'', thus triggering the STDP rule in both cases. This is further detailed in fig.\ref{fig:preciseTiming}, which shows which elements of the STDP rule are triggered by each spike in the network.

The precise spike times in this model have been chosen based on experimentation to ensure that the potentiation and the depression induced by the STDP rule on the plastic synapses have the same magnitude but opposite sign. Finally, the save neuron is used to allow the storage of the newly computed synaptic weight to memory, so that these can be retrieved at the end of the simulation.

Initially, synapses are set with a weight of zero, emphasizing that the output spike relies solely on the contribution of the teacher neuron. This underscores that the training process exclusively depends on the activity of the teacher neuron and the STDP learning rule.

\begin{figure}[tp]
\centering
\includegraphics[width=\linewidth]{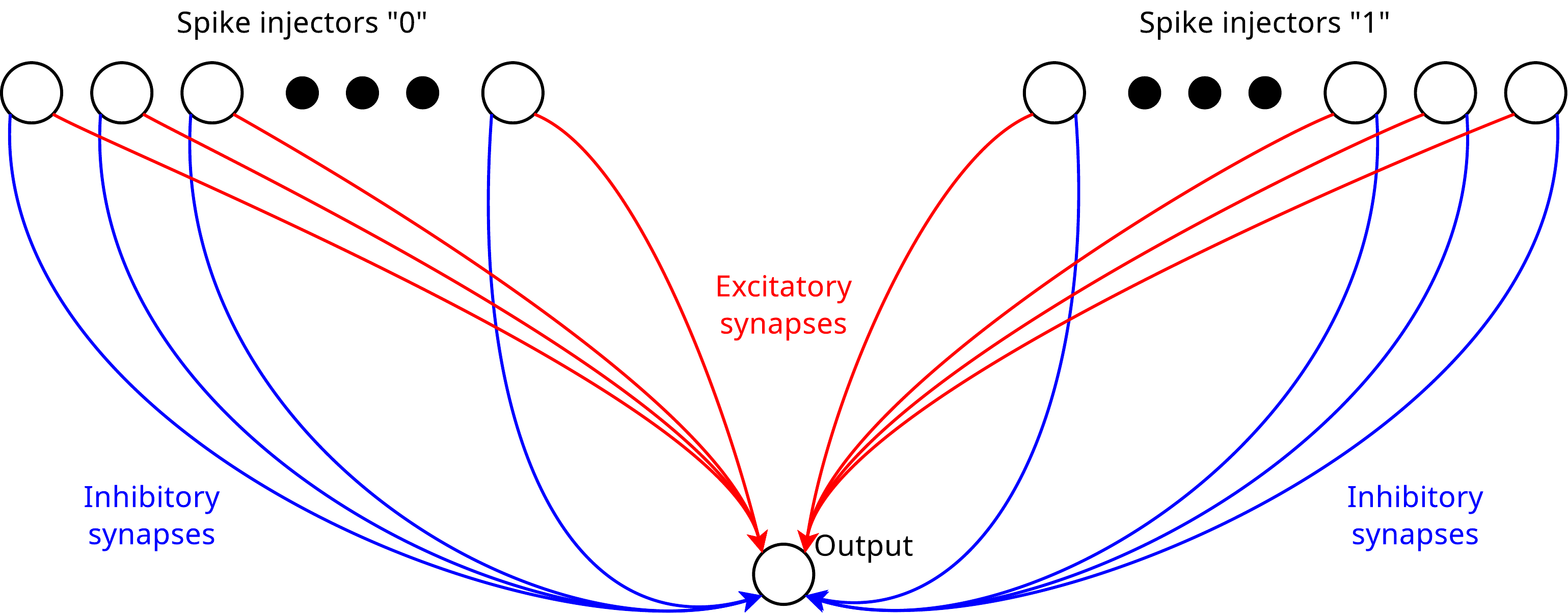}
\caption{The network used for testing.}
\label{fig:TestingNetwork}
\end{figure}

The synaptic weights obtained during this training phase are used in the testing network in fig.\ref{fig:TestingNetwork}.
The input pattern is injected in this network through the ``Spike Injectors 0'' and ``Spike Injectors 1'' populations.
In the first population a neuron fires if the corresponding bit of the input pattern is a ``0''.
On the contrary, if the bit is a ``1'', then the corresponding neuron of the ``Spike Injectors 1'' population fires.
In addition, all synapses are fixed, and the excitatory weights originating from ``Spike injector 0'' or ``Spike injector 1'' to the output neuron mirror the patterns learned by the synapses in the corresponding locations of the preceding network.
In contrast, the inhibitory weights stemming from ``Spike injector 0'' to the output neuron are guided by the weights learned by ``Spike injector 1'', and conversely, the inhibitory weights originating from ``Spike injector 1'' to the output neuron are influenced by the weights learned by ``Spike injector 0''.
Excitatory synapses from the ``Spike injector 1'' population and inhibitory synapses from ``Spike injector 0'' population have the same weight but opposite sign. The same applies to excitatory synapses from ``Spike injector 0'' and inhibitory synapses from ``Spike injector 1'' populations.

Following this pattern of connectivity:

\vspace{5pt}

\begin{center} 
$W^{I}_{SI0}(n) = -W^{E}_{SI1}(n)$ \\
\vspace{10pt}
$W^{I}_{SI1}(n) = -W^{E}_{SI0}(n)$
\end{center} 


Where:
\begin{itemize}[leftmargin=*]
    \item $\boldsymbol{W^{E}_{SI0}(n)}$ represents the weight of the excitatory synapse from the n-th neuron of the ``Spike injector 0'' population;
    \item $\boldsymbol{W^{I}_{SI0}(n)}$ represents the weight of the inhibitory synapse from the n-th neuron of the ``Spike injector 0'' population;
    \item $\boldsymbol{W^{E}_{SI1}(n)}$ represents the weight of the excitatory synapse from the n-th neuron of the ``Spike injector 1'' population;
    \item $\boldsymbol{W^{I}_{SI1}(n)}$ represents the weight of the inhibitory synapse from the n-th neuron of the ``Spike injector 1'' population;
\end{itemize}

This reciprocal relationship between excitatory and inhibitory weights ensures that the output neuron is able to select the pattern or patterns to respond to. Indeed, even in the case that the network is trained on multiple patterns, the output signal is generated always by the single output neuron present in the network.

In the cases where the network needs training on multiple patterns, each pattern will be trained separately, always starting from a network with plastic synaptic weights set to ``0''. The final value of the synaptic weights will be obtained by summing the weights resulting from training, synapse-by-synapse.
The weights obtained from the previous step are used as the basis for the homeostatic process.
During this step, after summing all the weights from the various training iterations, homeostasis is applied. This is modeled as a multiplier factor that re-scales the synaptic weights to obtain the minimum scaling value for each training pattern that allows the output neuron to fire exactly once for each learned pattern.
To complete this step, the summed and re-scaled synaptic weights are applied to the testing network in multiple iterations, adjusting the homeostatic factor at each iteration. Each training pattern is injected into this network, obtaining a homeostatic factor that may differ for each trained pattern.
Finally, the maximum among these factors is selected as the network homeostatic factor.

\renewcommand\tabularxcolumn[1]{m{#1}}

\begin{table}
\begin{tabularx}{\columnwidth}{m{1cm}@{}m{0.01cm}@{}m{2.5cm}@{}m{0.01cm}@{}X}
\hline
\multicolumn{1}{c}{\textbf{Name}} &
~ &
\multicolumn{1}{c}{\textbf{Variable}} &
~ &
\multicolumn{1}{c}{\textbf{Description}} \\
\hline
~ & ~ & ~ & ~ & ~ \\[-0.9em]
\hline

~ & ~ & ~ & ~ & ~ \\[-0.9em]
\centering\begin{tabular}[x]{@{}c@{}}\textit{Positives}\end{tabular}&
~ &
\centering\begin{tabular}[x]{@{}c@{}}$P_{i}$\end{tabular} &
~ &
Equals to $1$ in case the network emits a spike associated with the i-th input pattern\\
~ & ~ & ~ & ~ & ~ \\[-0.9em]
\hline

~ & ~ & ~ & ~ & ~ \\[-0.9em]
\centering\begin{tabular}[x]{@{}c@{}}\textit{Negatives}\end{tabular}&
~ &
\centering\begin{tabular}[x]{@{}c@{}}$N_{i}$\end{tabular} &
~ &
Equals to 1 in case the network does not emit a spike associated with the i-th input pattern\\
~ & ~ & ~ & ~ & ~ \\[-0.9em]
\hline

~ & ~ & ~ & ~ & ~ \\[-0.9em]
\centering\begin{tabular}[x]{@{}c@{}}\textit{True} \\ \textit{Positives}\end{tabular}&
~ &
\centering\begin{tabular}[x]{@{}c@{}}$t_{p}$\end{tabular} &
~ &
Number of spikes emitted associated with patterns learned by the network\\
~ & ~ & ~ & ~ & ~ \\[-0.9em]
\hline

~ & ~ & ~ & ~ & ~ \\[-0.9em]
\centering\begin{tabular}[x]{@{}c@{}}\textit{True} \\ \textit{Negatives}\end{tabular}&
~ &
\centering\begin{tabular}[x]{@{}c@{}}$t_{n}$\end{tabular} &
~ &
Number of patterns correctly not identified by the network\\
~ & ~ & ~ & ~ & ~ \\[-0.9em]
\hline

~ & ~ & ~ & ~ & ~ \\[-0.9em]
\centering\begin{tabular}[x]{@{}c@{}}\textit{False} \\ \textit{Positives}\end{tabular}&
~ &
\centering\begin{tabular}[x]{@{}c@{}}$f_{p}$\end{tabular} &
~ &
Number of spikes emitted, but not associated with patterns learned by the network\\
~ & ~ & ~ & ~ & ~ \\[-0.9em]
\hline

~ & ~ & ~ & ~ & ~ \\[-0.9em]
\centering\begin{tabular}[x]{@{}c@{}}\textit{False} \\ \textit{Negatives}\end{tabular}&
~ &
\centering\begin{tabular}[x]{@{}c@{}}$f_{n}$\end{tabular} &
~ &
Number of patterns on which the network was trained but that the network failed to identify\\
~ & ~ & ~ & ~ & ~ \\[-0.9em]
\hline
~ & ~ & ~ & ~ & ~ \\[-0.9em]

\multicolumn{1}{c}{\textbf{Name}} &
~ &
\multicolumn{1}{c}{\textbf{Formula}} &
~ &
\multicolumn{1}{c}{\textbf{Description}} \\
\hline
~ & ~ & ~ & ~ & ~ \\[-0.9em]
\hline

~ & ~ & ~ & ~ & ~ \\[-0.9em]
\centering\begin{tabular}[x]{@{}c@{}}\textit{Accuracy}\end{tabular} &
~ &
\centering \resizebox{0.8\hsize}{!}{~$\frac{t_{p}+t_{n}}{t_{p}+t_{n}+f_{p}+f_{n}}$} &
~ &
Proximity of the identification task to the training. It evaluates the overall performance of classification\\
~ & ~ & ~ & ~ & ~ \\[-0.9em]
\hline

~ & ~ & ~ & ~ & ~ \\[-0.9em]
\begin{tabular}[x]{@{}c@{}}\textit{Precision}\end{tabular}&
~ &
\centering \resizebox{0.4\hsize}{!}{$\frac{t_{p}}{t_{p}+f_{p}}$} &
~ &
Positive predicted value. This indicates the reliability of identification\\
~ & ~ & ~ & ~ & ~ \\[-0.9em]
\hline
~ & ~ & ~ & ~ & ~ \\[-0.9em]
\begin{tabular}[x]{@{}c@{}}\textit{Negative} \\ \textit{Prediction}\end{tabular}&
~ &
\centering \resizebox{0.4\hsize}{!}{$\frac{t_{n}}{t_{n}+f_{n}}$} &
~ &
Reliability of classification of distractions\\
~ & ~ & ~ & ~ & ~ \\[-0.9em]
\hline
~ & ~ & ~ & ~ & ~ \\[-0.9em]
\begin{tabular}[x]{@{}c@{}}\textit{Sensitivity}\end{tabular}&
~ &
\centering \resizebox{0.4\hsize}{!}{$\frac{t_{p}}{t_{p}+f_{n}}$} &
~ &
Focuses on how good is the performance in classifying attention\\
~ & ~ & ~ & ~ & ~ \\[-0.9em]
\hline
~ & ~ & ~ & ~ & ~ \\[-0.9em]
\begin{tabular}[x]{@{}c@{}}\textit{Specificity}\end{tabular}&
~ &
\centering \resizebox{0.4\hsize}{!}{$\frac{t_{n}}{t_{n}+f_{p}}$} &
~ &
Evaluates the performance in classifying distractions\\
~ & ~ & ~ & ~ & ~ \\[-0.9em]
\hline
\end{tabularx}
\caption{Classification metrics used for the evaluation of the performance of trained networks.}
\label{tab:Metrics}
\end{table}

The search for the re-scaling factor is performed initially through a binary search in the interval between the values of $0.0001$ and $1000$, reducing the interval until the values between the two extremes $x$ and $y$ is less than or equal to $0.0001$.
Then the search becomes linear in the interval $[x - 0.00001; y + 0.00001]$ with a step equal to $0.00001$, one order of magnitude smaller than the interval.
The precision of this step descends from the precision of the synaptic weights on SpiNNaker.
This precision is determined at runtime based upon the maximum weight value possible within the network \citep{vanAlbada2018PerformanceModel}.
For the network described here the minimum weight that can be represented is $2^{-11} \approx 0.0005$, so a linear search with a step size equal to $0.00001$ does not reduce the precision of the network.
Once the homeostatic factor is determined to allow the network to spike once for every pattern presented, the process moves to the testing and validation phase.
Because the homeostatic process re-scales all weights to obtain a specific required result, the original weights obtained through the STDP is of little relevance to the whole training process.

\subsection{Test and Validation phase}\label{label:TestingPhase}

During this phase, the objective is to validate the methodology for the training of a spatial feature classifier. To this end, we calculate the following classification metrics: Accuracy, Precision, Negative Prediction, Sensitivity, and Specificity \citep[as presented in~][]{Davies2021ANeuro-Robotics}.
These metrics are employed for class identification, as detailed in Table \ref{tab:Metrics}.

The network utilised is depicted in Figure \ref{fig:TestingNetwork}. All neurons operate as leaky integrate-and-fire units with delta synapses, as previously described.

\section{Results}\label{label:results}

\subsection{Single pattern training}\label{label:single_pattern}

To create a sufficiently broad testing space, the network in fig.\ref{fig:TestingNetwork} undergoes testing and validation using 10-bit patterns.
These patterns are represented by numbers in the range $[0;1023]$, where their binary representation effectively reflects the combination of spikes in the pattern.
The initial test focuses on the pattern expressed by the number $992_{10}=1111100000_{2}$ (the subscript numbers represent the base in which the code word is expressed).
Since the network comprises only 10 synapses per injector population, the details of the weights generated by the training step are documented fully in table \ref{tab:SinglePatternSynapticWeightsWithoutHomeostasis} to provide context for the discussion.

It is evident that the weights precisely mirror the pattern of ``0''s and ``1''s in the pattern.
Subsequently, homeostasis is applied to the group of synapses to ensure that the output neuron fires once when the pattern is presented to the network.
The resulting homeostasis factor is computed as $4.18817$, which re-scales the weight values to those presented in table \ref{tab:SinglePatternSynapticWeightsWithHomeostasis}.

With the weights outlined in the latter table, the network is validated using all possible combinations of spikes, showing that it produces only one output spike in response to the input $992_{10} = 1111100000_{2}$, in accordance with the training provided, demonstrating perfect pattern recognition, as shown in table \ref{tab:SinglePatternMetric}.

\begin{table}[!bp]
\noindent\resizebox{\columnwidth}{!}{
\begin{tabular}{l|llllllllll|}
\cline{2-11}
\multicolumn{1}{c|}{}                & \multicolumn{10}{c|}{Neuron ID} \\
\hline
\multicolumn{1}{|c|}{Population}     & \multicolumn{1}{c|}{9}       & \multicolumn{1}{c|}{8}       & \multicolumn{1}{c|}{7}       & \multicolumn{1}{c|}{6}       & \multicolumn{1}{c|}{5}       & \multicolumn{1}{c|}{4}       & \multicolumn{1}{c|}{3}       & \multicolumn{1}{c|}{2}       & \multicolumn{1}{c|}{1}       & \multicolumn{1}{c|}{0} \\
\hline
\multicolumn{1}{|l|}{Injector ``0''} & \multicolumn{1}{l|}{$0$}     & \multicolumn{1}{l|}{$0$}     & \multicolumn{1}{l|}{$0$}     & \multicolumn{1}{l|}{$0$}     & \multicolumn{1}{l|}{$0$}     & \multicolumn{1}{l|}{$0.367$} & \multicolumn{1}{l|}{$0.367$} & \multicolumn{1}{l|}{$0.367$} & \multicolumn{1}{l|}{$0.367$} & $0.367$                \\
\hline
\multicolumn{1}{|l|}{Injector ``1''} & \multicolumn{1}{l|}{$0.367$} & \multicolumn{1}{l|}{$0.367$} & \multicolumn{1}{l|}{$0.367$} & \multicolumn{1}{l|}{$0.367$} & \multicolumn{1}{l|}{$0.367$} & \multicolumn{1}{l|}{$0$}     & \multicolumn{1}{l|}{$0$}     & \multicolumn{1}{l|}{$0$}     & \multicolumn{1}{l|}{$0$}     & $0$                    \\
\hline
\end{tabular}
}
\caption{Trained synaptic weights for a single pattern expressed by the number $992_{10}=1111100000_{2}$}
\label{tab:SinglePatternSynapticWeightsWithoutHomeostasis}
\end{table}

\begin{table}[!h]
\noindent\resizebox{\columnwidth}{!}{
\begin{tabular}{l|llllllllll|}
\cline{2-11}
\multicolumn{1}{c|}{}                & \multicolumn{10}{c|}{Neuron ID} \\
\hline
\multicolumn{1}{|c|}{Population}     & \multicolumn{1}{c|}{9}       & \multicolumn{1}{c|}{8}       & \multicolumn{1}{c|}{7}       & \multicolumn{1}{c|}{6}       & \multicolumn{1}{c|}{5}       & \multicolumn{1}{c|}{4}       & \multicolumn{1}{c|}{3}       & \multicolumn{1}{c|}{2}       & \multicolumn{1}{c|}{1}       & \multicolumn{1}{c|}{0} \\
\hline
\multicolumn{1}{|l|}{Injector ``0''} & \multicolumn{1}{l|}{$0$}     & \multicolumn{1}{l|}{$0$}     & \multicolumn{1}{l|}{$0$}     & \multicolumn{1}{l|}{$0$}     & \multicolumn{1}{l|}{$0$}     & \multicolumn{1}{l|}{$1.538$} & \multicolumn{1}{l|}{$1.538$} & \multicolumn{1}{l|}{$1.538$} & \multicolumn{1}{l|}{$1.538$} & $1.538$                \\
\hline
\multicolumn{1}{|l|}{Injector ``1''} & \multicolumn{1}{l|}{$1.538$} & \multicolumn{1}{l|}{$1.538$} & \multicolumn{1}{l|}{$1.538$} & \multicolumn{1}{l|}{$1.538$} & \multicolumn{1}{l|}{$1.538$} & \multicolumn{1}{l|}{$0$}     & \multicolumn{1}{l|}{$0$}     & \multicolumn{1}{l|}{$0$}     & \multicolumn{1}{l|}{$0$}     & $0$                    \\
\hline
\end{tabular}
}
\caption{Re-scaled synaptic weights (after homeostasis)}
\label{tab:SinglePatternSynapticWeightsWithHomeostasis}
\end{table}

\begin{table}[bp]
\centering
\begin{tabular}{lll}
\hline
Metric              & Formula & Value                   \\ \hline
~&~\\[-0.9em] \hline

~&~\\[-0.9em]
Homeostatic value           & & $4.18817$                     \\
~&~\\[-0.9em] \hline
~&~\\[-0.9em] \hline

~&~\\[-0.9em]
Positives           & $P_{i}$ & $1$                     \\
~&~\\[-0.9em] \hline

~&~\\[-0.9em]
Negatives           & $N_{i}$ & $1023$                  \\
~&~\\[-0.9em] \hline

~&~\\[-0.9em]
True Positives      & $t_{p}$ & $1$                     \\
~&~\\[-0.9em] \hline

~&~\\[-0.9em]
True Negatives      & $t_{n}$ & $1023$                  \\
~&~\\[-0.9em] \hline

~&~\\[-0.9em]
False Positives     & $f_{p}$ & $0$                     \\
~&~\\[-0.9em] \hline

~&~\\[-0.9em]
False Negatives     & $f_{n}$ & $0$                     \\
~&~\\[-0.9em] \hline
~&~\\[-0.9em] \hline

~&~\\[-0.9em]
Accuracy            & \centering \resizebox{0.28\hsize}{!}{~$\frac{t_{p}+t_{n}}{t_{p}+t_{n}+f_{p}+f_{n}}$} & $\frac{1024}{1024}=1$   \\
~&~\\[-0.9em]\hline

~&~\\[-0.9em]
Precision           & \centering \resizebox{0.13\hsize}{!}{$\frac{t_{p}}{t_{p}+f_{p}}$} & $\frac{1}{1+0}=1$       \\
~&~\\[-0.9em]\hline

~&~\\[-0.9em]
\begin{tabular}[x]{@{}c@{}}Negative \\ Prediction\end{tabular} & \centering \resizebox{0.13\hsize}{!}{$\frac{t_{n}}{t_{n}+f_{n}}$} & $\frac{1023}{1023+0}=1$ \\
~&~\\[-0.9em]\hline

~&~\\[-0.9em]
Sensitivity         & \centering \resizebox{0.13\hsize}{!}{$\frac{t_{p}}{t_{p}+f_{n}}$} & $\frac{1}{1}=1$         \\
~&~\\[-0.9em]\hline

~&~\\[-0.9em]
Specificity         & \centering \resizebox{0.13\hsize}{!}{$\frac{t_{n}}{t_{n}+f_{p}}$} & $\frac{1023}{1023}=1$   \\
~&~\\[-0.9em]\hline
\end{tabular}
\caption{Classification metrics for the network trained with a single pattern.}
\label{tab:SinglePatternMetric}
\end{table}

\FloatBarrier

\subsection{Dual pattern training}\label{label:dual_pattern}

In addition to the single-pattern testing, a set of two-pattern training experiments has been conducted to investigate how training a single network on multiple patterns influences the recognition process.
Building upon the previous experiment, this set of experiments aims to elucidate how the disparity between the two learned patterns impacts the recognition process.
The first experiment aims to train the network to identify the patterns $992_{10}=1111100000_{2}$ and $960_{10}=1111000000_{2}$.
Between the two patterns there is only one bit difference in position $5$.
Network training takes this difference into account both at STDP training stage and at the homeostasis adjustment stage.
Indeed, the synaptic weights for the plastic synapses to the output neuron achieve the values presented in table \ref{tab:DualPatternTranedSynapticWeightsWithoutHomeostasis}.

\begin{table}[bp]
\noindent\resizebox{\columnwidth}{!}{
\begin{tabular}{l|llllllllll|}
\cline{2-11}
\multicolumn{1}{c|}{}                & \multicolumn{10}{c|}{Neuron ID} \\
\hline
\multicolumn{1}{|c|}{Population}     & \multicolumn{1}{c|}{9}       & \multicolumn{1}{c|}{8}       & \multicolumn{1}{c|}{7}       & \multicolumn{1}{c|}{6}       & \multicolumn{1}{c|}{5}       & \multicolumn{1}{c|}{4}       & \multicolumn{1}{c|}{3}       & \multicolumn{1}{c|}{2}       & \multicolumn{1}{c|}{1}       & \multicolumn{1}{c|}{0} \\
\hline
\multicolumn{1}{|l|}{Injector ``0''} & \multicolumn{1}{l|}{$0$}     & \multicolumn{1}{l|}{$0$}     & \multicolumn{1}{l|}{$0$}     & \multicolumn{1}{l|}{$0$}     & \multicolumn{1}{l|}{$0.367$}     & \multicolumn{1}{l|}{$0.734$} & \multicolumn{1}{l|}{$0.734$} & \multicolumn{1}{l|}{$0.734$} & \multicolumn{1}{l|}{$0.734$} & $0.734$                \\
\hline
\multicolumn{1}{|l|}{Injector ``1''} & \multicolumn{1}{l|}{$0.734$} & \multicolumn{1}{l|}{$0.734$} & \multicolumn{1}{l|}{$0.734$} & \multicolumn{1}{l|}{$0.734$} & \multicolumn{1}{l|}{$0.367$} & \multicolumn{1}{l|}{$0$}     & \multicolumn{1}{l|}{$0$}     & \multicolumn{1}{l|}{$0$}     & \multicolumn{1}{l|}{$0$}     & $0$                    \\
\hline
\end{tabular}
}
\caption{Trained synaptic weights for two patterns expressed by numbers $992_{10}=1111100000_{2}$ and $960_{10}=1111000000_{2}$}
    \label{tab:DualPatternTranedSynapticWeightsWithoutHomeostasis}
\end{table}

At a first glance, two main differences become evident when comparing these values with those related to the single-pattern experiment: in the first instance, weights of neurons $0$ to $4$ and $6$ to $9$ are doubled.
This is because the two-pattern experiment sums the corresponding synapses trained independently on the two patterns.
Therefore these synapses are reinforced twice, and their weights are doubled.

\begin{table*}[tp]
\centering
\resizebox{\linewidth}{!}{
\begin{tabular}{@{}lccccccccccccccc@{}}
\hline
Pattern 1 &
$992_{10}$ &
$992_{10}$ &
$992_{10}$ &
$992_{10}$ &
$992_{10}$ &
$992_{10}$ &
$992_{10}$ &
$992_{10}$ &
$992_{10}$ &
$992_{10}$ &
$992_{10}$ &
$992_{10}$ &
$992_{10}$ &
$992_{10}$ \\ \hline

Pattern 2 &
$1008_{10}$ &
$1016_{10}$ &
$1020_{10}$ &
$1022_{10}$ &
$1023_{10}$ &
$960_{10}$ &
$896_{10}$ &
$768_{10}$ &
$512_{10}$ &
$0_{10}$ &
$16_{10}$ &
$24_{10}$ &
$28_{10}$ &
$30_{10}$ \\ \hline

\begin{tabular}[x]{@{}l@{}}Hamming \\ Distance\end{tabular}&
$1$ &
$2$ &
$3$ &
$4$ &
$5$ &
$1$ &
$2$ &
$3$ &
$4$ &
$5$ &
$6$ &
$7$ &
$8$ &
$9$ \\ \hline

\begin{tabular}[x]{@{}l@{}}Homeostasis \\ factor\end{tabular}&
$ 2.3265$   &
$ 2.6177$   &
$ 2.9914$   &
$ 3.4907$   &
$ 4.1875$   &
$ 2.3265$   &
$ 2.6177$   &
$ 2.9914$   &
$ 3.4907$   &
$ 4.1875$   &
$ 5.2354$   &
$ 6.9814$   &
$10.4708$   &
$20.9415$   \\ \hline

~&~&~&~&~&~&~&~&~&~&~&~&~&~&~\\[-0.9em] \hline

~&~\\[-0.9em]
Positives &
$2$ & 
$4$ & 
$8$ & 
$16$ & 
$32$ & 
$2$ & 
$4$ & 
$8$ & 
$16$ & 
$32$ & 
$64$ & 
$128$ & 
$256$ & 
$512$ \\
~&~\\[-0.9em] \hline

~&~\\[-0.9em]
Negatives &
$1022$ & 
$1020$ & 
$1016$ & 
$1008$ & 
$992$ & 
$1022$ & 
$1020$ & 
$1016$ & 
$1008$ & 
$992$ & 
$960$ & 
$896$ & 
$768$ & 
$512$ \\
~&~\\[-0.9em] \hline

~&~\\[-0.9em]
\begin{tabular}[x]{@{}l@{}}True \\ Positives\end{tabular}&
$2$ & 
$2$ & 
$2$ & 
$2$ & 
$2$ & 
$2$ & 
$2$ & 
$2$ & 
$2$ & 
$2$ & 
$2$ & 
$2$ & 
$2$ & 
$2$ \\
~&~\\[-0.9em] \hline

~&~\\[-0.9em]
\begin{tabular}[x]{@{}l@{}}True \\ Negatives\end{tabular}&
$1022$ & 
$1020$ & 
$1016$ & 
$1008$ & 
$992$ & 
$1022$ & 
$1020$ & 
$1016$ & 
$1008$ & 
$992$ & 
$960$ & 
$896$ & 
$768$ & 
$512$ \\
~&~\\[-0.9em] \hline

~&~\\[-0.9em]
\begin{tabular}[x]{@{}l@{}}False \\ Positives\end{tabular}&
$0$ & 
$2$ & 
$6$ & 
$14$ & 
$30$ & 
$0$ & 
$2$ & 
$6$ & 
$14$ & 
$30$ & 
$62$ & 
$126$ & 
$254$ & 
$510$ \\
~&~\\[-0.9em] \hline

~&~\\[-0.9em]
\begin{tabular}[x]{@{}l@{}}False \\ Negatives\end{tabular}&
$0$ & 
$0$ & 
$0$ & 
$0$ & 
$0$ & 
$0$ & 
$0$ & 
$0$ & 
$0$ & 
$0$ & 
$0$ & 
$0$ & 
$0$ & 
$0$ \\
~&~\\[-0.9em] \hline

~&~&~&~&~&~&~&~&~&~&~&~&~&~&~\\[-0.9em] \hline

~&~\\[-0.9em]
Accuracy &
$1$ & 
$0.998$ & 
$0.994$ & 
$0.986$ & 
$0.971$ & 
$1$ & 
$0.998$ & 
$0.994$ & 
$0.986$ & 
$0.971$ & 
$0.939$ & 
$0.877$ & 
$0.752$ & 
$0.502$ \\
~&~\\[-0.9em]\hline

~&~\\[-0.9em]
Precision &
$1$ & 
$0.5$ & 
$0.25$ & 
$0.125$ & 
$0.0625$ & 
$1$ & 
$0.5$ & 
$0.25$ & 
$0.125$ & 
$0.0625$ & 
$0.03125$ & 
$0.0156$ & 
$0.00781$ & 
$0.00391$ \\
~&~\\[-0.9em]\hline

~&~\\[-0.9em]
\begin{tabular}[x]{@{}l@{}}Negative \\ Prediction\end{tabular} &
$1$ & 
$1$ & 
$1$ & 
$1$ & 
$1$ & 
$1$ & 
$1$ & 
$1$ & 
$1$ & 
$1$ & 
$1$ & 
$1$ & 
$1$ & 
$1$ \\
~&~\\[-0.9em]\hline

~&~\\[-0.9em]
Sensitivity &
$1$ & 
$1$ & 
$1$ & 
$1$ & 
$1$ & 
$1$ & 
$1$ & 
$1$ & 
$1$ & 
$1$ & 
$1$ & 
$1$ & 
$1$ & 
$1$ \\
~&~\\[-0.9em]\hline

~&~\\[-0.9em]
Specificity &
$1$ &
$0.998$ &
$0.994$ &
$0.986$ &
$0.971$ &
$1$ &
$0.998$ &
$0.994$ &
$0.986$ &
$0.971$ &
$0.939$ &
$0.877$ &
$0.751$ &
$0.501$ \\
~&~\\[-0.9em]\hline
\end{tabular}
}
\caption{Classification metrics for the network trained with two patterns}
\label{tab:DualPatternMetric}
\end{table*}

In the second instance, it is possible to notice that the weights of the synapses from neuron $5$ is evenly distributed between the two injector populations.
This distribution arises from the training process: while the pattern $992_{10}=1111100000_{2}$ trained the network to detect a $1$ in position $5$, the pattern $960_{10}=1111000000_{2}$ trained the network to detect a $0$ in the same position.

This means that neither of the neurons with ID $5$ contributes to the generation of the output spike.
In the testing network (fig.\ref{fig:TestingNetwork}) both the inhibitory and excitatory synapses from both injector populations have the same weight:

\vspace{-15pt}

\noindent
\begin{equation} \label{eq:NullWeight1}
I_{5} = W_{\text{exc5}|\text{Inj0}} - W_{\text{inh5}|\text{Inj0}} + \\ W_{\text{exc5}|\text{Inj1}} - W_{\text{inh5}|\text{Inj1}} = 
\end{equation}

\vspace{-25pt}

\noindent
\begin{equation} \label{eq:NullWeight2}
= \underbrace{0.367 - 0.367}_\text{\normalsize\begin{tabular}[x]{@{}c@{}}for a ``0'' \\ spike\end{tabular}} + \underbrace{0.367 - 0.367}_\text{\normalsize\begin{tabular}[x]{@{}c@{}}for a ``1'' \\ spike\end{tabular}} = 0
\end{equation}

\vspace{-5pt}

Therefore, the input current to the output neuron contributed by neuron $5$ in either spike injector population is null.
The homeostasis process takes this into account by increasing the overall value of the other contributing neurons by the same amount of the missing synapses.
In fact, the homeostatic parameter in this instance is

\vspace{-15pt}

\noindent
\begin{equation}
2.32647 \approx \frac{4.18817}{2} \times \frac{10}{9}
\end{equation}
\vspace{-15pt}

Where the value $4.18817$ is the homeostatic value from the single pattern experiment, and $\nicefrac{10}{9}$ represents the fact that one of the synapses is not contributing to the identification, and therefore all the other synapses need to be stronger.

In these conditions, it is possible to evaluate the network performance metrics in the detection of the patterns by testing all the possible combinations.
In this case the network positively identifies only the two trained patterns ($992_{10}=1111100000_{2}$ and $960_{10}=1111000000_{2}$) behaving as the perfect classifier.

As we test the network with patterns increasingly divergent from the original $992_{10}=1111100000_{2}$ pattern, several general trends become evident:
\begin{itemize}
    \item The synapses related to the bits that are different among the two patterns have weights evenly distributed between the two injector populations. In this way, the output neuron does not depend on these inputs, which can be considered ``don't care'' synapses or bits.
    \item The homeostatic factor increases with the number of ``don't care'' bits, to account for the fewer synapses that contribute to the detection.
    \item The network's performance metrics demonstrate a noticeable deterioration as the number of dissimilarities between the patterns learned by the network increases. This degradation in performance underscores the sensitivity of the network to discrepancies among the learned patterns.
    \item The deterioration in the performance of the pattern recognition task is related to the Hamming Distance \citep{Hamming1950ErrorCodes,Tomlinson2017Error-CorrectionDecoding} between the two learned patterns: the number of ``don't care'' bits in the network and therefore the number of patterns that the network is able to identify.
\end{itemize}

These results can be seen in table \ref{tab:DualPatternMetric}. The position of the ``don't care'' bits or synapses in the pattern is irrelevant for the purpose of this task. This can be seen in the cases where the Hamming Distance is 1 to 5: even though the patterns to learn are different, the classification metrics are equal and dependent only on the Hamming Distance between them.

A special mention should be made for the case of two patterns that are completely opposite.
In our case $992_{10}=1111100000_{2}$ and $31_{10}=0000011111_{2}$ have a Hamming Distance of $10$.
In this scenario, none of the synapses in the network would contribute to the identification of the pattern, and therefore, the output neuron would never spike.
Consequently, the homeostatic factor becomes infinite ($+\infty$).
As a result, all the classification metrics become incalculable since no output spike is generated under any circumstances, and therefore this combination of patterns is not included as a result in the table.

\subsection{Multiple pattern training}\label{label:multi_pattern}

\begin{table}[]
\noindent
\resizebox{\columnwidth}{!}{%
\begin{tabular}{lcccccccccc}
\multicolumn{11}{l}{~}\\
\cline{2-11}
\multicolumn{1}{l|}{} 
& \multicolumn{10}{c|}{Population ``0''}
\\ \hline
\multicolumn{1}{|l|}{Neuron ID}                                                                           & \multicolumn{1}{c|}{10} & \multicolumn{1}{c|}{9} & \multicolumn{1}{c|}{8} & \multicolumn{1}{c|}{7} & \multicolumn{1}{c|}{6} & \multicolumn{1}{c|}{5} & \multicolumn{1}{c|}{4} & \multicolumn{1}{c|}{3} & \multicolumn{1}{c|}{2} & \multicolumn{1}{c|}{1} \\ \hline
\multicolumn{1}{|l|}{\begin{tabular}[c]{@{}l@{}}Code Word ``$0$''\\ Unit Synaptic Contributions\end{tabular}} & \multicolumn{1}{c|}{1}  & \multicolumn{1}{c|}{1} & \multicolumn{1}{c|}{1} & \multicolumn{1}{c|}{1} & \multicolumn{1}{c|}{1} & \multicolumn{1}{c|}{1} & \multicolumn{1}{c|}{1} & \multicolumn{1}{c|}{1} & \multicolumn{1}{c|}{1} & \multicolumn{1}{c|}{1} \\ \hline
\multicolumn{1}{|l|}{\begin{tabular}[c]{@{}l@{}}Code word ``$1$''\\ Unit Synaptic Contribution\end{tabular}}  & \multicolumn{1}{c|}{1}  & \multicolumn{1}{c|}{1} & \multicolumn{1}{c|}{1} & \multicolumn{1}{c|}{1} & \multicolumn{1}{c|}{1} & \multicolumn{1}{c|}{1} & \multicolumn{1}{c|}{1} & \multicolumn{1}{c|}{1} & \multicolumn{1}{c|}{1} & \multicolumn{1}{c|}{0} \\ \hline
\multicolumn{1}{|l|}{\begin{tabular}[c]{@{}l@{}}Code word ``$2$''\\ Unit Synaptic Contribution\end{tabular}}  & \multicolumn{1}{c|}{1}  & \multicolumn{1}{c|}{1} & \multicolumn{1}{c|}{1} & \multicolumn{1}{c|}{1} & \multicolumn{1}{c|}{1} & \multicolumn{1}{c|}{1} & \multicolumn{1}{c|}{1} & \multicolumn{1}{c|}{1} & \multicolumn{1}{c|}{0} & \multicolumn{1}{c|}{1} \\ \hline

\multicolumn{1}{|l|}{~}  & \multicolumn{1}{c|}{~}  & \multicolumn{1}{c|}{~} & \multicolumn{1}{c|}{~} & \multicolumn{1}{c|}{~} & \multicolumn{1}{c|}{~} & \multicolumn{1}{c|}{~} & \multicolumn{1}{c|}{~} & \multicolumn{1}{c|}{~} & \multicolumn{1}{c|}{~} & \multicolumn{1}{c|}{~} \\ [-0.9em] \hline 

\multicolumn{1}{|l|}{Final Unit Synaptic Weights}                                                         & \multicolumn{1}{c|}{\textcolor{blue}{3}}  & \multicolumn{1}{c|}{\textcolor{blue}{3}} & \multicolumn{1}{c|}{\textcolor{blue}{3}} & \multicolumn{1}{c|}{\textcolor{blue}{3}} & \multicolumn{1}{c|}{\textcolor{blue}{3}} & \multicolumn{1}{c|}{\textcolor{blue}{3}} & \multicolumn{1}{c|}{\textcolor{blue}{3}} & \multicolumn{1}{c|}{\textcolor{blue}{3}} & \multicolumn{1}{c|}{\textcolor{blue}{2}} & \multicolumn{1}{c|}{\textcolor{blue}{2}} \\ \hline

\multicolumn{11}{l}{}   \\ [-0.5em]

\multicolumn{11}{l}{~}\\

\cline{2-11} 
\multicolumn{1}{l|}{}                                                                                     & \multicolumn{10}{c|}{Population ``1''}
\\ \hline
\multicolumn{1}{|l|}{Neuron ID}                                                                           & \multicolumn{1}{c|}{10} & \multicolumn{1}{c|}{9} & \multicolumn{1}{c|}{8} & \multicolumn{1}{c|}{7} & \multicolumn{1}{c|}{6} & \multicolumn{1}{c|}{5} & \multicolumn{1}{c|}{4} & \multicolumn{1}{c|}{3} & \multicolumn{1}{c|}{2} & \multicolumn{1}{c|}{1} \\ \hline
\multicolumn{1}{|l|}{\begin{tabular}[c]{@{}l@{}}Code Word ``$0$''\\ Unit Synaptic Contributions\end{tabular}} & \multicolumn{1}{c|}{0}  & \multicolumn{1}{c|}{0} & \multicolumn{1}{c|}{0} & \multicolumn{1}{c|}{0} & \multicolumn{1}{c|}{0} & \multicolumn{1}{c|}{0} & \multicolumn{1}{c|}{0} & \multicolumn{1}{c|}{0} & \multicolumn{1}{c|}{0} & \multicolumn{1}{c|}{0} \\ \hline
\multicolumn{1}{|l|}{\begin{tabular}[c]{@{}l@{}}Code word ``$1$''\\ Unit Synaptic Contribution\end{tabular}}  & \multicolumn{1}{c|}{0}  & \multicolumn{1}{c|}{0} & \multicolumn{1}{c|}{0} & \multicolumn{1}{c|}{0} & \multicolumn{1}{c|}{0} & \multicolumn{1}{c|}{0} & \multicolumn{1}{c|}{0} & \multicolumn{1}{c|}{0} & \multicolumn{1}{c|}{0} & \multicolumn{1}{c|}{1} \\ \hline
\multicolumn{1}{|l|}{\begin{tabular}[c]{@{}l@{}}Code word ``$2$''\\ Unit Synaptic Contribution\end{tabular}}  & \multicolumn{1}{c|}{0}  & \multicolumn{1}{c|}{0} & \multicolumn{1}{c|}{0} & \multicolumn{1}{c|}{0} & \multicolumn{1}{c|}{0} & \multicolumn{1}{c|}{0} & \multicolumn{1}{c|}{0} & \multicolumn{1}{c|}{0} & \multicolumn{1}{c|}{1} & \multicolumn{1}{c|}{0} \\ \hline

\multicolumn{1}{|l|}{~}  & \multicolumn{1}{c|}{~}  & \multicolumn{1}{c|}{~} & \multicolumn{1}{c|}{~} & \multicolumn{1}{c|}{~} & \multicolumn{1}{c|}{~} & \multicolumn{1}{c|}{~} & \multicolumn{1}{c|}{~} & \multicolumn{1}{c|}{~} & \multicolumn{1}{c|}{~} & \multicolumn{1}{c|}{~} \\ [-0.9em] \hline

\multicolumn{1}{|l|}{Final Unit Synaptic Weights}                                                         & \multicolumn{1}{c|}{\textcolor{red}{0}}  & \multicolumn{1}{c|}{\textcolor{red}{0}} & \multicolumn{1}{c|}{\textcolor{red}{0}} & \multicolumn{1}{c|}{\textcolor{red}{0}} & \multicolumn{1}{c|}{\textcolor{red}{0}} & \multicolumn{1}{c|}{\textcolor{red}{0}} & \multicolumn{1}{c|}{\textcolor{red}{0}} & \multicolumn{1}{c|}{\textcolor{red}{0}} & \multicolumn{1}{c|}{\textcolor{red}{1}} & \multicolumn{1}{c|}{\textcolor{red}{1}} \\ \hline
\multicolumn{11}{l}{~}\\
\end{tabular}%
}

    \caption{Trained unit synaptic weights for the three code words ``$0$'', ``$1$'' and ``$2$''}
    \label{tab:ExampleThreeCodeWordsTrainedUnitSynaticWeights}
\end{table}

\begin{table}[]
\resizebox{\columnwidth}{!}{%
\begin{tabular}{|c|c|c|}
\multicolumn{3}{c}{~} \\
\hline
\begin{tabular}[c]{@{}c@{}}Code\\ Word\end{tabular} &
\begin{tabular}[c]{@{}c@{}}Synaptic\\ Contribution\end{tabular} &
\begin{tabular}[c]{@{}c@{}}Unit\\ Weight\end{tabular} \\
\hline

``$0$'' &
\begin{tabular}[c]{@{}c@{}}$\textcolor{blue}{3}+\textcolor{blue}{3}+\textcolor{blue}{3}+\textcolor{blue}{3}+\textcolor{blue}{3}+\textcolor{blue}{3}+\textcolor{blue}{3}+\textcolor{blue}{3}+\textcolor{blue}{2}+\textcolor{blue}{2}-$\\$(\textcolor{red}{0}+\textcolor{red}{0}+\textcolor{red}{0}+\textcolor{red}{0}+\textcolor{red}{0}+\textcolor{red}{0}+\textcolor{red}{0}+\textcolor{red}{0}+\textcolor{red}{1}+\textcolor{red}{1})$\end{tabular} &
$26$ \\
\hline

``$1$'' &
\begin{tabular}[c]{@{}c@{}}$\textcolor{blue}{3}+\textcolor{blue}{3}+\textcolor{blue}{3}+\textcolor{blue}{3}+\textcolor{blue}{3}+\textcolor{blue}{3}+\textcolor{blue}{3}+\textcolor{blue}{3}+\textcolor{blue}{2}+\textcolor{red}{1}-$\\$(\textcolor{red}{0}+\textcolor{red}{0}+\textcolor{red}{0}+\textcolor{red}{0}+\textcolor{red}{0}+\textcolor{red}{0}+\textcolor{red}{0}+\textcolor{red}{0}+\textcolor{red}{1}+\textcolor{blue}{2})$\end{tabular} &
$24$ \\
\hline

``$2$'' &
\begin{tabular}[c]{@{}c@{}}$\textcolor{blue}{3}+\textcolor{blue}{3}+\textcolor{blue}{3}+\textcolor{blue}{3}+\textcolor{blue}{3}+\textcolor{blue}{3}+\textcolor{blue}{3}+\textcolor{blue}{3}+\textcolor{red}{1}+\textcolor{blue}{2}-$\\$(\textcolor{red}{0}+\textcolor{red}{0}+\textcolor{red}{0}+\textcolor{red}{0}+\textcolor{red}{0}+\textcolor{red}{0}+\textcolor{red}{0}+\textcolor{red}{0}+\textcolor{blue}{2}+\textcolor{red}{1})$\end{tabular} &
$24$ \\
\hline

\multicolumn{3}{l}{}   \\ [-0.4em]

\end{tabular}%
}
\caption{Example of synaptic unit contributions for the three code words ``$0$'', ``$1$'' and ``$2$'', colored numbers indicate the source of the corresponding weight from table \ref{tab:ExampleThreeCodeWordsTrainedUnitSynaticWeights}}
\label{tab:ExampleUnitContributionsThreeCodeWords}
\end{table}

Finally, the network was tested with three code words. As discussed in the previous case, the accuracy of the network relies on the similarities between the various code words, and in particular on the Hamming Distance across all code words trained.

To ensure the test was conducted with the largest possible set comprising all 10-bit code words, the Hamming Distance between all possible combinations of 10-bit numbers was computed, and only the first occurrence of the code words for each Hamming Distance was recorded, where the code words were all different. The resulting combinations of code words is described in table \ref{tab:ThreeCodeWordsUnitWeights}.

The network was trained using the same protocol as in previous experiments, this time using three code words. The weights obtained in this way are then re-scaled simulating an homeostatic process so that, where possible, the network would spike for all three trained patterns.
However, as shown in table \ref{tab:ThreeCodeWordsUnitWeights}, not in all cases this is possible: with specific combinations of code words it is not possible to have the positive identification of one or even two code words.

To explain this behaviour we can start considering each training iteration as providing a synaptic weight unit contribution to the ``0'' population or to the ``1'' population. When the code word is then re-applied during homeostasis, these unit weights contribute to the final weight applied to the output neuron. If the sum of excitatory and inhibitory unit weights is positive, then the homeostasis finds the smallest factor for which the network fires for all the code words. In alternative, one or two code words are ``discarded'' during this process and will appear in the ``False Negative'' count.

For example, in the case of code words ``$0$'', ``$1$'' and ``$2$'', the final weight unit contribution to the output neuron is described in table \ref{tab:ExampleThreeCodeWordsTrainedUnitSynaticWeights}:

As for the code word $0$ all the neurons in population ``0'' are spiking, these neuron contribute to the output neuron by exciting it through the excitatory ``0'' synapses and inhibit it through the inhibitory ``1'' synapses (see fig.\ref{fig:TestingNetwork}). In this case the final contribution is positive and includes $26$ unit weights. If we repeat this process for all the code words in the example, the synaptic weights contributions are as described in table \ref{tab:ExampleUnitContributionsThreeCodeWords}, where colored numbers indicate the source of the corresponding weight from table \ref{tab:ExampleThreeCodeWordsTrainedUnitSynaticWeights}.

As all the contributions are positive, in this case applying the appropriate homeostatic factor to the weights will lead to the output neuron firing (at least) for the trained code words. Considering the STDP parameters applied to the network, one synaptic weight unit is equal to $0.3671875$ and the homeostatic factor required for the network is estimated in this case equal to $1.74513$.

The table describing all the combinations of code words, their Hamming Distance, the synaptic unit weights and homeostasis factors computed for this case is presented in table \ref{tab:ThreeCodeWordsUnitWeights}.

Table \ref{tab:ThreeCodeWordsMetrics} introduces the performance metrics of the test and validation network, and clearly illustrates two trends: maintaining two code words constant while progressively increasing the Hamming Distance of the remaining one from the others results in the deterioration of synaptic unit weights. These weights reach zero or become negative, rendering one or more code word(s) no longer positively identifiable by the network.

\begin{table*}[]
\resizebox{0.89\textwidth}{!}{%
\begin{tabular}{|c|c|c|@{}c@{}|c|c|c|@{}c@{}|c|c|c|@{}c@{}|c|}
\hline
CW1 &
CW2 &
CW3 &
~ &
\begin{tabular}[c]{@{}c@{}}HD\\ (1,2)\end{tabular} &
\begin{tabular}[c]{@{}c@{}}HD\\ (1,3)\end{tabular} &
\begin{tabular}[c]{@{}c@{}}HD\\ (2,3)\end{tabular} &
~ &
\begin{tabular}[c]{@{}c@{}}Code Word 1\\ Unit Weight\end{tabular} &
\begin{tabular}[c]{@{}c@{}}Code Word 2\\ Unit Weight\end{tabular} &
\begin{tabular}[c]{@{}c@{}}Code Word 3\\ Unit Weight\end{tabular} &
~ &
\begin{tabular}[c]{@{}c@{}}Homeostatic\\ factor\end{tabular} \\ \hline
0 &
1 &
2 &
&
1 &
1 &
2 &
&
26 &
24 &
24 &
&
1.74513 \\ \hline

0 &
1 &
6 &
&
1 &
2 &
3 &
&
24 &
22 &
20 &
&
2.09442 \\ \hline
0 &
1 &
14 &
&
1 &
3 &
4 &
&
22 &
20 &
16 &
&
2.61791 \\ \hline
0 &
1 &
30 &
&
1 &
4 &
5 &
&
20 &
18 &
12 &
&
3.49203 \\ \hline
0 &
1 &
62 &
&
1 &
5 &
6 &
&
18 &
16 &
8 &
&
5.23937 \\ \hline
0 &
1 &
126 &
&
1 &
6 &
7 &
&
16 &
14 &
4 &
&
10.47873 \\ \hline
0 &
1 &
254 &
&
1 &
7 &
8 &
&
14 &
12 &
\multicolumn{1}{>{\columncolor[rgb]{1,0.5,0.5}[0pt]}c|}{0} &
&
3.4907 \\ \hline
0 &
1 &
510 &
&
1 &
8 &
9 &
&
12 &
10 &
\multicolumn{1}{>{\columncolor[rgb]{1,0.5,0.5}[0pt]}c|}{-4} &
&
4.19016 \\ \hline
0 &
1 &
1022 &
&
1 &
9 &
10 &
&
10 &
8 &
\multicolumn{1}{>{\columncolor[rgb]{1,0.5,0.5}[0pt]}c|}{-8} &
&
5.23804 \\ \hline
0 &
3 &
5 &
&
2 &
2 &
2 &
&
22 &
22 &
22 &
&
1.90382 \\ \hline
0 &
3 &
12 &
&
2 &
2 &
4 &
&
22 &
18 &
18 &
&
2.32669 \\ \hline
0 &
3 &
13 &
&
2 &
3 &
3 &
&
20 &
20 &
18 &
&
2.32713 \\ \hline
0 &
3 &
28 &
&
2 &
3 &
5 &
&
20 &
16 &
14 &
&
2.99203 \\ \hline
0 &
3 &
29 &
&
2 &
4 &
4 &
&
18 &
18 &
14 &
&
2.99203 \\ \hline
0 &
3 &
60 &
&
2 &
4 &
6 &
&
18 &
14 &
10 &
&
4.18883 \\ \hline
0 &
3 &
61 &
&
2 &
5 &
5 &
&
16 &
16 &
10 &
&
4.18618 \\ \hline
0 &
3 &
124 &
&
2 &
5 &
7 &
&
16 &
12 &
6 &
&
6.98405 \\ \hline
0 &
3 &
125 &
&
2 &
6 &
6 &
&
14 &
14 &
6 &
&
6.98405 \\ \hline
0 &
3 &
252 &
&
2 &
6 &
8 &
&
14 &
10 &
2 &
&
21 \\ \hline
0 &
3 &
253 &
&
2 &
7 &
7 &
&
12 &
12 &
2 &
&
21 \\ \hline
0 &
3 &
508 &
&
2 &
7 &
9 &
&
12 &
8 &
\multicolumn{1}{>{\columncolor[rgb]{1,0.5,0.5}[0pt]}c|}{-2} &
&
5.23671 \\ \hline
0 &
3 &
509 &
&
2 &
8 &
8 &
&
10 &
10 &
\multicolumn{1}{>{\columncolor[rgb]{1,0.5,0.5}[0pt]}c|}{-2} &
&
4.18883 \\ \hline
0 &
3 &
1020 &
&
2 &
8 &
10 &
&
10 &
6 &
\multicolumn{1}{>{\columncolor[rgb]{1,0.5,0.5}[0pt]}c|}{-6} &
&
6.98139 \\ \hline
0 &
3 &
1021 &
&
2 &
9 &
9 &
&
8 &
8 &
\multicolumn{1}{>{\columncolor[rgb]{1,0.5,0.5}[0pt]}c|}{-6} &
&
5.23671 \\ \hline
0 &
7 &
25 &
&
3 &
3 &
4 &
&
18 &
16 &
16 &
&
2.61791 \\ \hline
0 &
7 &
56 &
&
3 &
3 &
6 &
&
18 &
12 &
12 &
&
3.49025 \\ \hline
0 &
7 &
57 &
&
3 &
4 &
5 &
&
16 &
14 &
12 &
&
3.49025 \\ \hline
0 &
7 &
120 &
&
3 &
4 &
7 &
&
16 &
10 &
8 &
&
5.23582 \\ \hline
0 &
7 &
121 &
&
3 &
5 &
6 &
&
14 &
12 &
8 &
&
5.23582 \\ \hline
0 &
7 &
248 &
&
3 &
5 &
8 &
&
14 &
8 &
4 &
&
10.47873 \\ \hline
0 &
7 &
249 &
&
3 &
6 &
7 &
&
12 &
10 &
4 &
&
10.47873 \\ \hline
0 &
7 &
504 &
&
3 &
6 &
9 &
&
12 &
6 &
\multicolumn{1}{>{\columncolor[rgb]{1,0.5,0.5}[0pt]}c|}{0} &
&
6.98139 \\ \hline
0 &
7 &
505 &
&
3 &
7 &
8 &
&
10 &
8 &
\multicolumn{1}{>{\columncolor[rgb]{1,0.5,0.5}[0pt]}c|}{0} &
&
5.23671 \\ \hline
0 &
7 &
1016 &
&
3 &
7 &
10 &
&
10 &
4 &
\multicolumn{1}{>{\columncolor[rgb]{1,0.5,0.5}[0pt]}c|}{-4} &
&
10.47607 \\ \hline
0 &
7 &
1017 &
&
3 &
8 &
9 &
&
8 &
6 &
\multicolumn{1}{>{\columncolor[rgb]{1,0.5,0.5}[0pt]}c|}{-4} &
&
6.98139 \\ \hline
0 &
15 &
51 &
&
4 &
4 &
4 &
&
14 &
14 &
14 &
&
2.99203 \\ \hline
0 &
15 &
113 &
&
4 &
4 &
6 &
&
14 &
10 &
10 &
&
4.18972 \\ \hline
0 &
15 &
115 &
&
4 &
5 &
5 &
&
12 &
12 &
10 &
&
4.18972 \\ \hline
0 &
15 &
240 &
&
4 &
4 &
8 &
&
14 &
6 &
6 &
&
6.9805 \\ \hline
0 &
15 &
241 &
&
4 &
5 &
7 &
&
12 &
8 &
6 &
&
6.9805 \\ \hline
0 &
15 &
243 &
&
4 &
6 &
6 &
&
10 &
10 &
6 &
&
6.9805 \\ \hline
0 &
15 &
496 &
&
4 &
5 &
9 &
&
12 &
4 &
2 &
&
20.95745 \\ \hline
0 &
15 &
497 &
&
4 &
6 &
8 &
&
10 &
6 &
2 &
&
20.95745 \\ \hline
0 &
15 &
499 &
&
4 &
7 &
7 &
&
8 &
8 &
2 &
&
20.95745 \\ \hline
0 &
15 &
1008 &
&
4 &
6 &
10 &
&
10 &
2 &
\multicolumn{1}{>{\columncolor[rgb]{1,0.5,0.5}[0pt]}c|}{-2} &
&
20.95213 \\ \hline
0 &
15 &
1009 &
&
4 &
7 &
9 &
&
8 &
4 &
\multicolumn{1}{>{\columncolor[rgb]{1,0.5,0.5}[0pt]}c|}{-2} &
&
10.47341 \\ \hline
0 &
15 &
1011 &
&
4 &
8 &
8 &
&
6 &
6 &
\multicolumn{1}{>{\columncolor[rgb]{1,0.5,0.5}[0pt]}c|}{-2} &
&
6.98139 \\ \hline
0 &
31 &
227 &
&
5 &
5 &
6 &
&
10 &
8 &
8 &
&
5.23582 \\ \hline
0 &
31 &
481 &
&
5 &
5 &
8 &
&
10 &
4 &
4 &
&
10.47164 \\ \hline
0 &
31 &
483 &
&
5 &
6 &
7 &
&
8 &
6 &
4 &
&
10.47164 \\ \hline
0 &
31 &
992 &
&
5 &
5 &
10 &
&
10 &
\multicolumn{1}{>{\columncolor[rgb]{1,0.5,0.5}[0pt]}c|}{0} &
\multicolumn{1}{>{\columncolor[rgb]{1,0.5,0.5}[0pt]}c|}{0} &
&
4.19016 \\ \hline
0 &
31 &
993 &
&
5 &
6 &
9 &
&
8 &
2 &
\multicolumn{1}{>{\columncolor[rgb]{1,0.5,0.5}[0pt]}c|}{0} &
&
20.94681 \\ \hline
0 &
31 &
995 &
&
5 &
7 &
8 &
&
6 &
4 &
\multicolumn{1}{>{\columncolor[rgb]{1,0.5,0.5}[0pt]}c|}{0} &
&
10.47341 \\ \hline
0 &
63 &
455 &
&
6 &
6 &
6 &
&
6 &
6 &
6 &
&
6.98139 \\ \hline
0 &
63 &
963 &
&
6 &
6 &
8 &
&
6 &
2 &
2 &
&
20.94681 \\ \hline
0 &
63 &
967 &
&
6 &
7 &
7 &
&
4 &
4 &
2 &
&
20.94681 \\ \hline
\end{tabular}%
}
\caption{Combinations of code words (CW1, CW2 and CW3) trained on the network, their Hamming Distance HD(x,y), the synaptic unit weight, computed as described in the text, and the resulting homeostatic factor.}
\label{tab:ThreeCodeWordsUnitWeights}
\end{table*}

\begin{table*}[]
\resizebox{0.92\textwidth}{!}{%
\begin{tabular}{|c|c|c|@{}l@{}|c|c|c|c|@{}l@{}|c|c|c|c|c|}
\hline
CW1 &
  CW2 &
  CW3 &
  ~ &
  \begin{tabular}[c]{@{}c@{}}True\\ Positives\end{tabular} &
  \begin{tabular}[c]{@{}c@{}}True\\ Negatives\end{tabular} &
  \begin{tabular}[c]{@{}c@{}}False\\ Positives\end{tabular} &
  \begin{tabular}[c]{@{}c@{}}False\\ Negatives\end{tabular} &
  ~ &
  \multicolumn{1}{c|}{Accuracy} &
  \multicolumn{1}{c|}{Precision} &
  \multicolumn{1}{c|}{\begin{tabular}[c]{@{}c@{}}Negative\\ Prediction\end{tabular}} &
  \multicolumn{1}{c|}{Sensitivity} &
  \multicolumn{1}{c|}{Specificity} \\ \hline
0 & 1  & 2    &  & 3 & 1021 & 0   & 0                                                          &  & 1.000 & 1.000 & 1.000 & 1.000 & 1.000 \\ \hline
0 & 1  & 6    &  & 3 & 1017 & 4   & 0                                                          &  & 0.996 & 0.429 & 1.000 & 1.000 & 0.996 \\ \hline
0 & 1  & 14   &  & 3 & 1003 & 18  & 0                                                          &  & 0.982 & 0.143 & 1.000 & 1.000 & 0.982 \\ \hline
0 & 1  & 30   &  & 3 & 963  & 58  & 0                                                          &  & 0.943 & 0.049 & 1.000 & 1.000 & 0.943 \\ \hline
0 & 1  & 62   &  & 3 & 873  & 148 & 0                                                          &  & 0.855 & 0.020 & 1.000 & 1.000 & 0.855 \\ \hline
0 & 1  & 126  &  & 3 & 705  & 316 & 0                                                          &  & 0.691 & 0.009 & 1.000 & 1.000 & 0.690 \\ \hline
0 & 1  & 254  &  & 2 & 1014 & 7   & \multicolumn{1}{>{\columncolor[rgb]{1,0.5,0.5}[0pt]}c|}{1} &  & 0.992 & 0.222 & 0.999 & 0.667 & 0.993 \\ \hline
0 & 1  & 510  &  & 2 & 1013 & 8   & \multicolumn{1}{>{\columncolor[rgb]{1,0.5,0.5}[0pt]}c|}{1} &  & 0.991 & 0.200 & 0.999 & 0.667 & 0.992 \\ \hline
0 & 1  & 1022 &  & 2 & 1012 & 9   & \multicolumn{1}{>{\columncolor[rgb]{1,0.5,0.5}[0pt]}c|}{1} &  & 0.990 & 0.182 & 0.999 & 0.667 & 0.991 \\ \hline
0 & 3  & 5    &  & 3 & 1020 & 1   & 0                                                          &  & 0.999 & 0.750 & 1.000 & 1.000 & 0.999 \\ \hline
0 & 3  & 12   &  & 3 & 1013 & 8   & 0                                                          &  & 0.992 & 0.273 & 1.000 & 1.000 & 0.992 \\ \hline
0 & 3  & 13   &  & 3 & 1013 & 8   & 0                                                          &  & 0.992 & 0.273 & 1.000 & 1.000 & 0.992 \\ \hline
0 & 3  & 28   &  & 3 & 998  & 23  & 0                                                          &  & 0.978 & 0.115 & 1.000 & 1.000 & 0.977 \\ \hline
0 & 3  & 29   &  & 3 & 998  & 23  & 0                                                          &  & 0.978 & 0.115 & 1.000 & 1.000 & 0.977 \\ \hline
0 & 3  & 60   &  & 3 & 963  & 58  & 0                                                          &  & 0.943 & 0.049 & 1.000 & 1.000 & 0.943 \\ \hline
0 & 3  & 61   &  & 3 & 963  & 58  & 0                                                          &  & 0.943 & 0.049 & 1.000 & 1.000 & 0.943 \\ \hline
0 & 3  & 124  &  & 3 & 817  & 204 & 0                                                          &  & 0.801 & 0.014 & 1.000 & 1.000 & 0.800 \\ \hline
0 & 3  & 125  &  & 3 & 817  & 204 & 0                                                          &  & 0.801 & 0.014 & 1.000 & 1.000 & 0.800 \\ \hline
0 & 3  & 252  &  & 3 & 591  & 430 & 0                                                          &  & 0.580 & 0.007 & 1.000 & 1.000 & 0.579 \\ \hline
0 & 3  & 253  &  & 3 & 591  & 430 & 0                                                          &  & 0.580 & 0.007 & 1.000 & 1.000 & 0.579 \\ \hline
0 & 3  & 508  &  & 2 & 977  & 44  & \multicolumn{1}{>{\columncolor[rgb]{1,0.5,0.5}[0pt]}c|}{1} &  & 0.956 & 0.043 & 0.999 & 0.667 & 0.957 \\ \hline
0 & 3  & 509  &  & 2 & 1013 & 8   & \multicolumn{1}{>{\columncolor[rgb]{1,0.5,0.5}[0pt]}c|}{1} &  & 0.991 & 0.200 & 0.999 & 0.667 & 0.992 \\ \hline
0 & 3  & 1020 &  & 2 & 967  & 54  & \multicolumn{1}{>{\columncolor[rgb]{1,0.5,0.5}[0pt]}c|}{1} &  & 0.946 & 0.036 & 0.999 & 0.667 & 0.947 \\ \hline
0 & 3  & 1021 &  & 2 & 1012 & 9   & \multicolumn{1}{>{\columncolor[rgb]{1,0.5,0.5}[0pt]}c|}{1} &  & 0.990 & 0.182 & 0.999 & 0.667 & 0.991 \\ \hline
0 & 7  & 25   &  & 3 & 1008 & 13  & 0                                                          &  & 0.987 & 0.188 & 1.000 & 1.000 & 0.987 \\ \hline
0 & 7  & 56   &  & 3 & 982  & 39  & 0                                                          &  & 0.962 & 0.071 & 1.000 & 1.000 & 0.962 \\ \hline
0 & 7  & 57   &  & 3 & 982  & 39  & 0                                                          &  & 0.962 & 0.071 & 1.000 & 1.000 & 0.962 \\ \hline
0 & 7  & 120  &  & 3 & 922  & 99  & 0                                                          &  & 0.903 & 0.029 & 1.000 & 1.000 & 0.903 \\ \hline
0 & 7  & 121  &  & 3 & 922  & 99  & 0                                                          &  & 0.903 & 0.029 & 1.000 & 1.000 & 0.903 \\ \hline
0 & 7  & 248  &  & 3 & 787  & 234 & 0                                                          &  & 0.771 & 0.013 & 1.000 & 1.000 & 0.771 \\ \hline
0 & 7  & 249  &  & 3 & 787  & 234 & 0                                                          &  & 0.771 & 0.013 & 1.000 & 1.000 & 0.771 \\ \hline
0 & 7  & 504  &  & 2 & 892  & 129 & \multicolumn{1}{>{\columncolor[rgb]{1,0.5,0.5}[0pt]}c|}{1} &  & 0.873 & 0.015 & 0.999 & 0.667 & 0.874 \\ \hline
0 & 7  & 505  &  & 2 & 977  & 44  & \multicolumn{1}{>{\columncolor[rgb]{1,0.5,0.5}[0pt]}c|}{1} &  & 0.956 & 0.043 & 0.999 & 0.667 & 0.957 \\ \hline
0 & 7  & 1016 &  & 2 & 847  & 174 & \multicolumn{1}{>{\columncolor[rgb]{1,0.5,0.5}[0pt]}c|}{1} &  & 0.829 & 0.011 & 0.999 & 0.667 & 0.830 \\ \hline
0 & 7  & 1017 &  & 2 & 967  & 54  & \multicolumn{1}{>{\columncolor[rgb]{1,0.5,0.5}[0pt]}c|}{1} &  & 0.946 & 0.036 & 0.999 & 0.667 & 0.947 \\ \hline
0 & 15 & 51   &  & 3 & 1002 & 19  & 0                                                          &  & 0.981 & 0.136 & 1.000 & 1.000 & 0.981 \\ \hline
0 & 15 & 113  &  & 3 & 957  & 64  & 0                                                          &  & 0.938 & 0.045 & 1.000 & 1.000 & 0.937 \\ \hline
0 & 15 & 115  &  & 3 & 957  & 64  & 0                                                          &  & 0.938 & 0.045 & 1.000 & 1.000 & 0.937 \\ \hline
0 & 15 & 240  &  & 3 & 859  & 162 & 0                                                          &  & 0.842 & 0.018 & 1.000 & 1.000 & 0.841 \\ \hline
0 & 15 & 241  &  & 3 & 859  & 162 & 0                                                          &  & 0.842 & 0.018 & 1.000 & 1.000 & 0.841 \\ \hline
0 & 15 & 243  &  & 3 & 859  & 162 & 0                                                          &  & 0.842 & 0.018 & 1.000 & 1.000 & 0.841 \\ \hline
0 & 15 & 496  &  & 3 & 632  & 389 & 0                                                          &  & 0.620 & 0.008 & 1.000 & 1.000 & 0.619 \\ \hline
0 & 15 & 497  &  & 3 & 632  & 389 & 0                                                          &  & 0.620 & 0.008 & 1.000 & 1.000 & 0.619 \\ \hline
0 & 15 & 499  &  & 3 & 632  & 389 & 0                                                          &  & 0.620 & 0.008 & 1.000 & 1.000 & 0.619 \\ \hline
0 & 15 & 1008 &  & 2 & 637  & 384 & \multicolumn{1}{>{\columncolor[rgb]{1,0.5,0.5}[0pt]}c|}{1} &  & 0.624 & 0.005 & 0.998 & 0.667 & 0.624 \\ \hline
0 & 15 & 1009 &  & 2 & 847  & 174 & \multicolumn{1}{>{\columncolor[rgb]{1,0.5,0.5}[0pt]}c|}{1} &  & 0.829 & 0.011 & 0.999 & 0.667 & 0.830 \\ \hline
0 & 15 & 1011 &  & 2 & 967  & 54  & \multicolumn{1}{>{\columncolor[rgb]{1,0.5,0.5}[0pt]}c|}{1} &  & 0.946 & 0.036 & 0.999 & 0.667 & 0.947 \\ \hline
0 & 31 & 227  &  & 3 & 931  & 90  & 0                                                          &  & 0.912 & 0.032 & 1.000 & 1.000 & 0.912 \\ \hline
0 & 31 & 481  &  & 3 & 767  & 254 & 0                                                          &  & 0.752 & 0.012 & 1.000 & 1.000 & 0.751 \\ \hline
0 & 31 & 483  &  & 3 & 767  & 254 & 0                                                          &  & 0.752 & 0.012 & 1.000 & 1.000 & 0.751 \\ \hline
0 & 31 & 992  &  & 1 & 1021 & 0   & \multicolumn{1}{>{\columncolor[rgb]{1,0.5,0.5}[0pt]}c|}{2} &  & 0.998 & 1.000 & 0.998 & 0.333 & 1.000 \\ \hline
0 & 31 & 993  &  & 2 & 637  & 384 & \multicolumn{1}{>{\columncolor[rgb]{1,0.5,0.5}[0pt]}c|}{1} &  & 0.624 & 0.005 & 0.998 & 0.667 & 0.624 \\ \hline
0 & 31 & 995  &  & 2 & 847  & 174 & \multicolumn{1}{>{\columncolor[rgb]{1,0.5,0.5}[0pt]}c|}{1} &  & 0.829 & 0.011 & 0.999 & 0.667 & 0.830 \\ \hline
0 & 63 & 455  &  & 3 & 893  & 128 & 0                                                          &  & 0.875 & 0.023 & 1.000 & 1.000 & 0.875 \\ \hline
0 & 63 & 963  &  & 3 & 638  & 383 & 0                                                          &  & 0.626 & 0.008 & 1.000 & 1.000 & 0.625 \\ \hline
0 & 63 & 967  &  & 3 & 638  & 383 & 0                                                          &  & 0.626 & 0.008 & 1.000 & 1.000 & 0.625 \\ \hline
\end{tabular}%
}
\caption{Combinations of code words (CW1, CW2 and CW3) trained on the network, and the corresponding test classification metrics obtained from simulations.}
\label{tab:ThreeCodeWordsMetrics}
\end{table*}

Simultaneously, as the synaptic unit weight decreases, the homeostatic factor increases to compensate for the limited efficacy of the incoming excitation. This holds true until the network is no longer able to detect one of the code words, which is then automatically excluded from the identification task during the search for a valid homeostatic factor: indeed the neural network cannot emit a spike in case the output neuron excitation results in a ``$0$'' or even a negative number. The unit weights for which one or two code words are no longer identifiable are highlighted in red in table \ref{tab:ThreeCodeWordsUnitWeights}.

\begin{figure*}
    \begin{subfigure}{0.32\textwidth}
        \centering
        \includegraphics[trim={48pt 32pt 15pt 25pt},clip,width=1\textwidth]{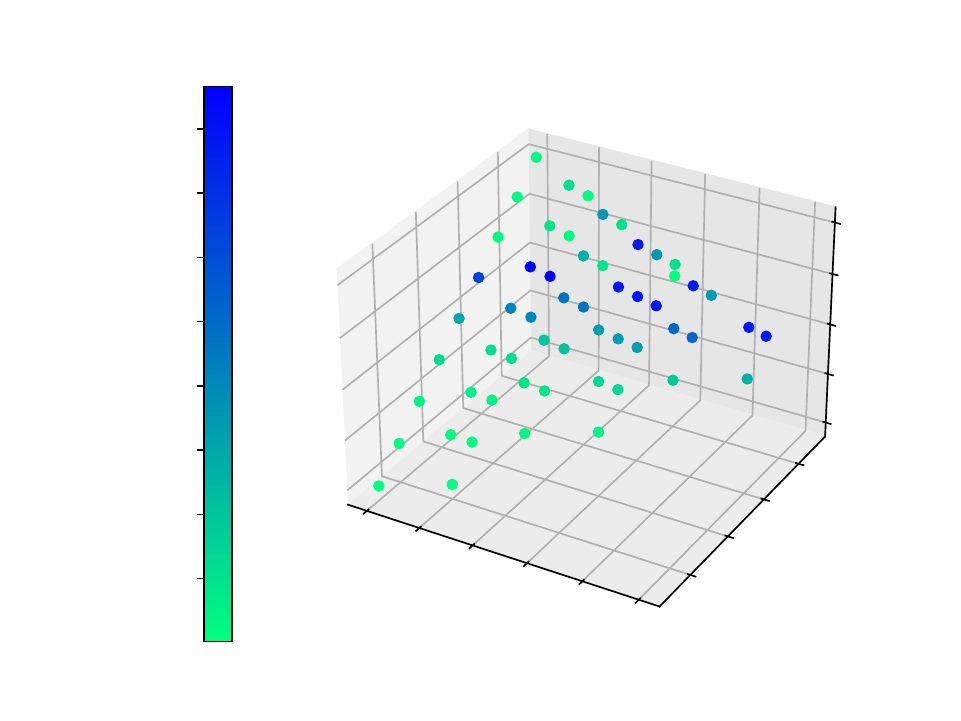}
        \caption{True positive + False Positives}
    \end{subfigure}
    \begin{subfigure}{0.32\textwidth}
        \centering
        \includegraphics[trim={48pt 32pt 15pt 25pt},clip,width=1\textwidth]{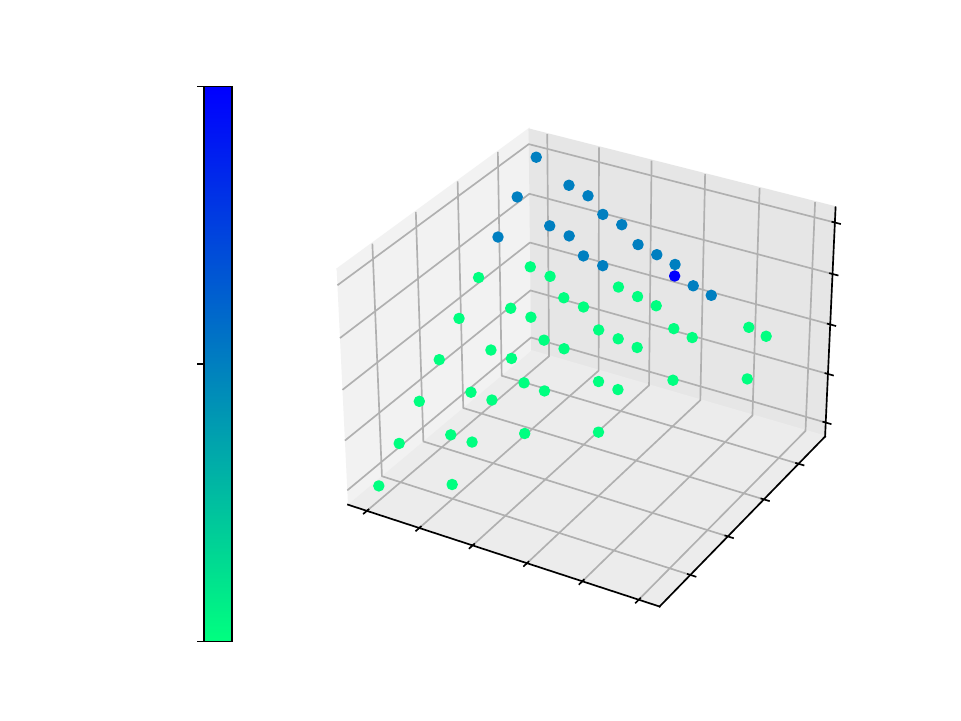}
        \caption{False negatives}
    \end{subfigure}
    \begin{subfigure}{0.32\textwidth}
        \centering
        \includegraphics[trim={48pt 32pt 15pt 25pt},clip,width=1\textwidth]{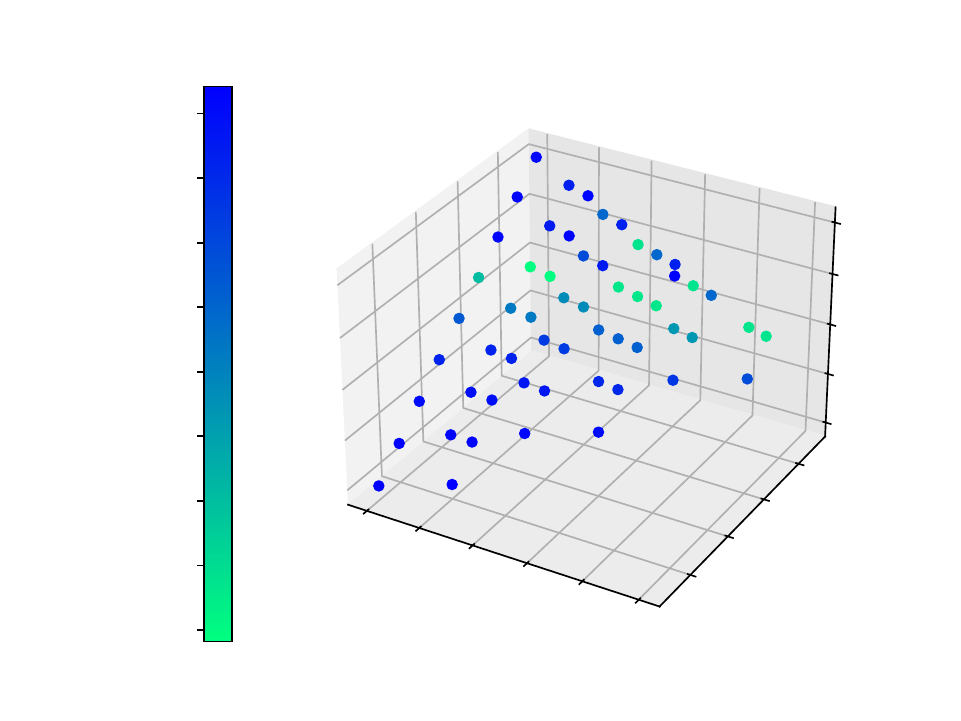}
        \caption{True negatives}
    \end{subfigure}
    \hfill
    \begin{subfigure}{0.32\textwidth}
        \centering
        \includegraphics[trim={48pt 32pt 15pt 25pt},clip,width=1\textwidth]{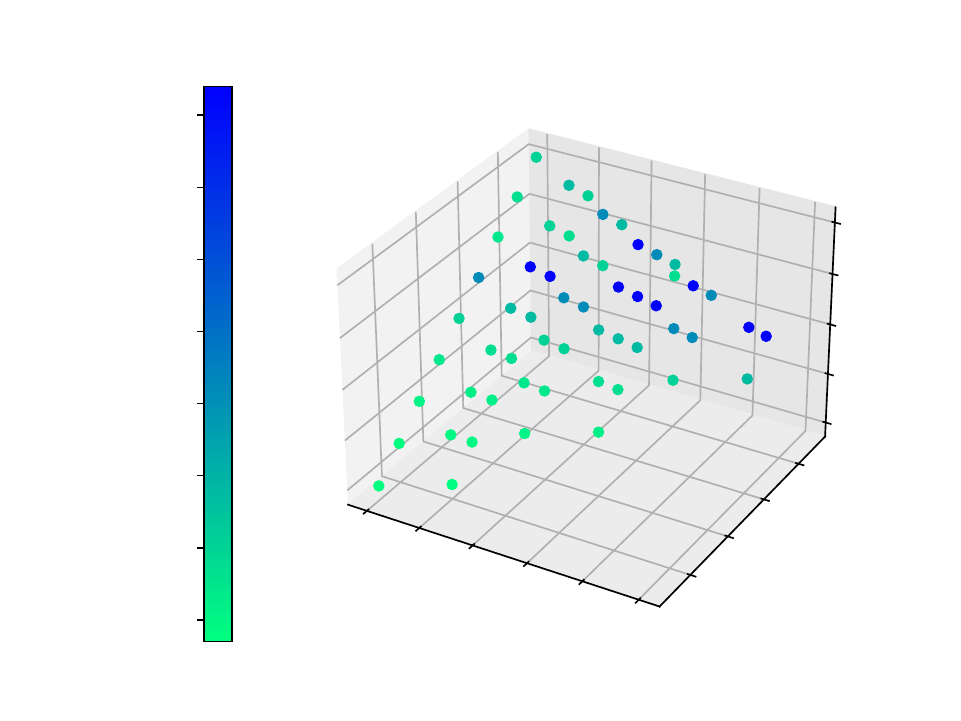}
        \caption{Homeostatic factor}
    \end{subfigure}
    \begin{subfigure}{0.32\textwidth}
        \centering
        \includegraphics[trim={48pt 32pt 15pt 25pt},clip,width=1\textwidth]{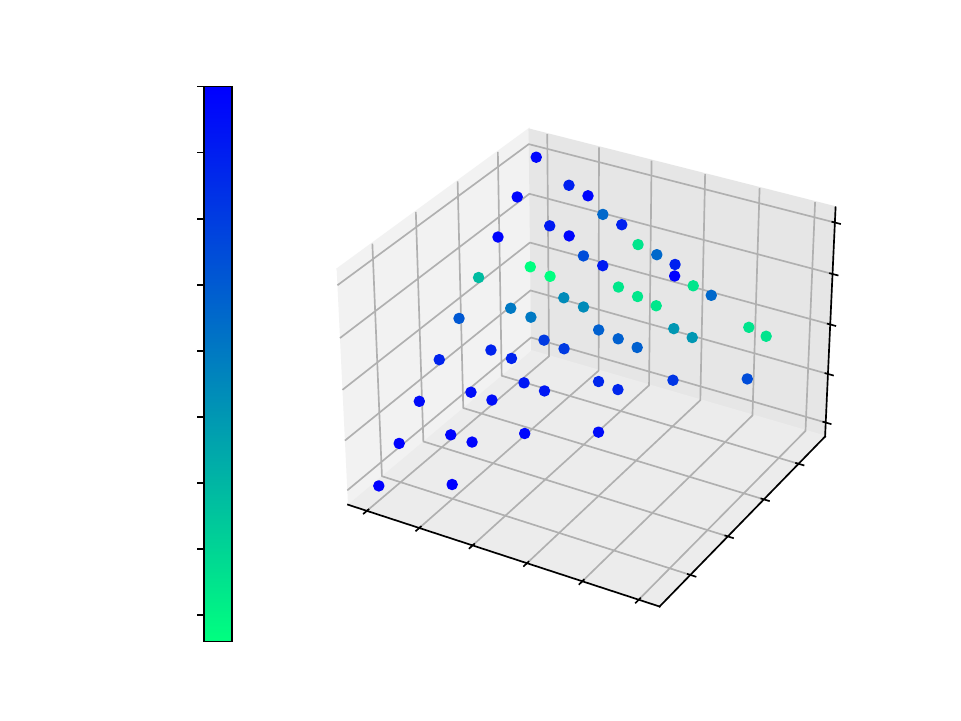}
        \caption{Accuracy}
    \end{subfigure}
    \begin{subfigure}{0.32\textwidth}
        \centering
        \includegraphics[trim={48pt 32pt 15pt 25pt},clip,width=1\textwidth]{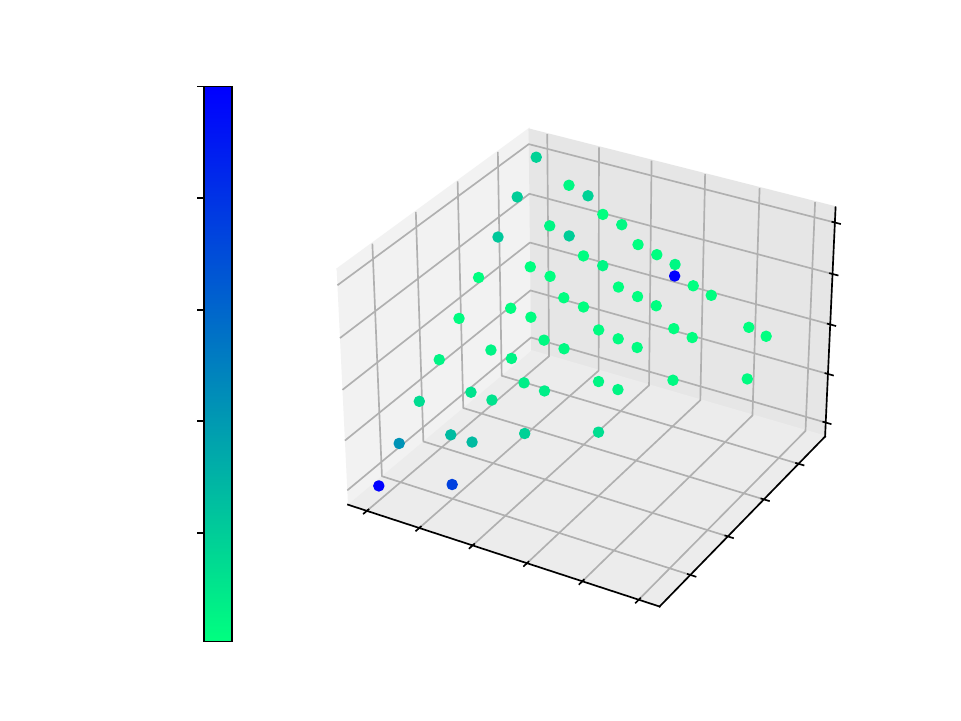}
        \caption{Precision}
    \end{subfigure}
    \hfill
    \begin{subfigure}{0.32\textwidth}
        \centering
        \includegraphics[trim={48pt 32pt 15pt 25pt},clip,width=1\textwidth]{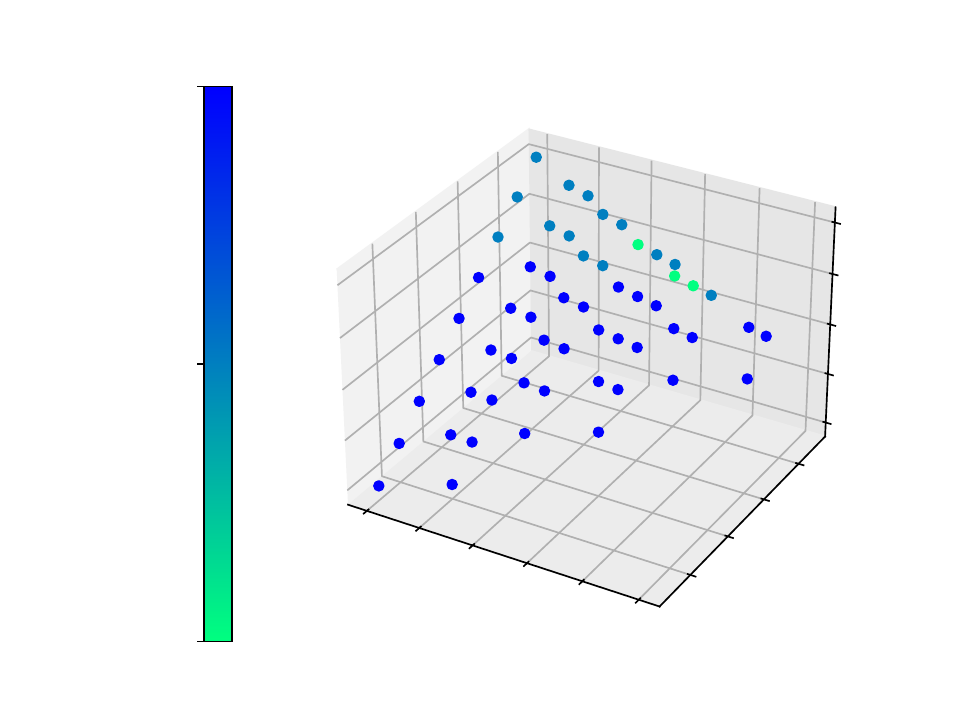}
        \caption{Negative prediction}
    \end{subfigure}
    \begin{subfigure}{0.32\textwidth}
        \centering
        \includegraphics[trim={48pt 32pt 15pt 25pt},clip,width=1\textwidth]{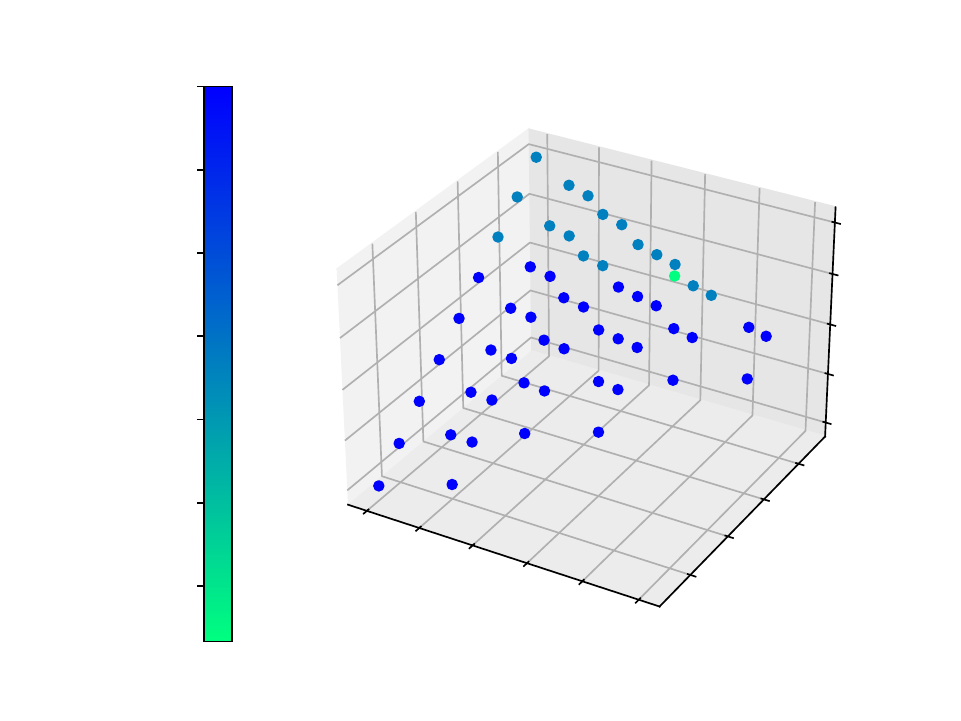}
        \caption{Sensitivity}
    \end{subfigure}
    \begin{subfigure}{0.32\textwidth}
        \centering
        \includegraphics[trim={48pt 32pt 15pt 25pt},clip,width=1\textwidth]{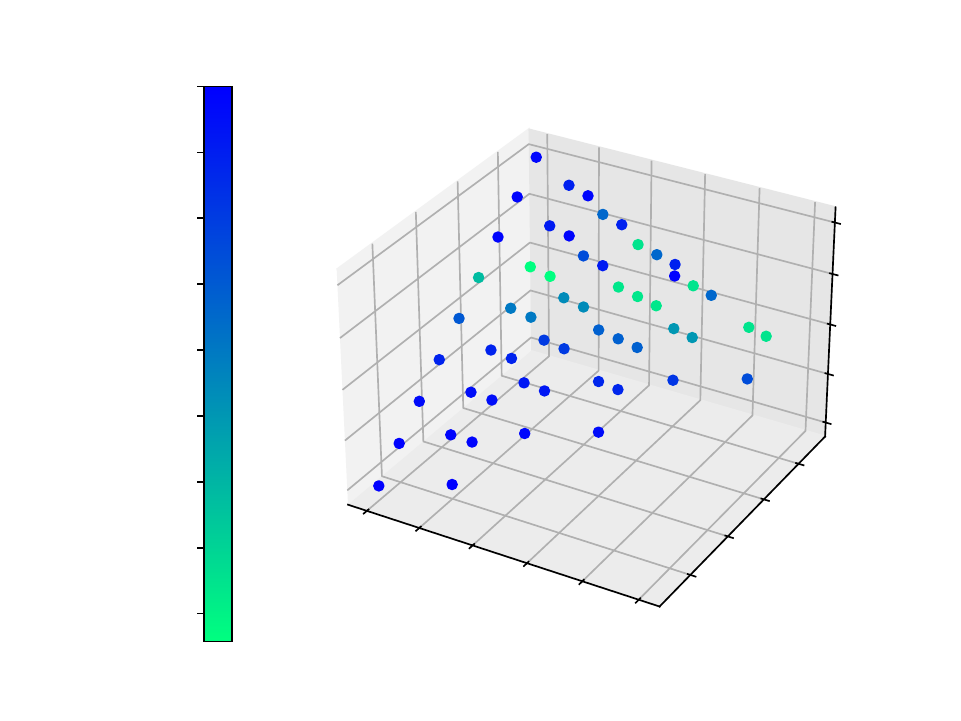}
        \caption{Specificity}
    \end{subfigure}
    \caption{Statistical classification metrics for the multiple code words training. The axes represent the Hamming Distance between the various code words as indicated in each graph.}
    \label{fig:Metrics}
\end{figure*}

Comparing table \ref{tab:ThreeCodeWordsUnitWeights} with table \ref{tab:ThreeCodeWordsMetrics}, it is possible to notice that the false negatives appear in correspondence to the $0$ or negative unit weights. These classification metrics show that the accuracy is linked to the Hamming Distance between code words. However, when one code word is dropped from identification, the negative prediction and specificity parameters receive lower values, but the overall accuracy improves, as the number of false positive identifications drastically reduces. However, the precision parameter decreases rapidly as the number of false positives increases.

The classification metrics presented in table \ref{tab:ThreeCodeWordsMetrics} are also displayed graphically in fig.\ref{fig:Metrics}. In these graphs it is possible to notice how the closer the experiments are to the bottom left corner, the better the classification metrics that represent the outcome. Indeed, the bottom left corner represents experiments using three code words whose Hamming Distance among them is minimum. In these graphs the three axis HD(x,y) indicate the Hamming Distance between the ``x'' and ``y'' code words.

In particular it is possible to notice also how there is an inverse relation between the homeostatic factor and the overall accuracy of the identification task. As the distance between code words increases, the number of synapses with overall contribution ``0'' (``don't care'' synapses) or small (``care little'' synapses) increases, and these require a higher homeostatic factor to allow the remaining synapses to trigger an output spike. However, this also causes the number of false positive identifications to increase, which in turn reduces the overall accuracy of the identification task of the network. 

\section{Conclusions}\label{label:conclusions}

This paper presented a method for training a spiking neural network to identify spatial patterns. Validation results show that a spiking network trained this way is able to successfully complete this task. The testing involved training on a single pattern, two patterns and three patterns, covering  all meaningful combinations of code words. In these experiments, it became evident that specific parameter values for the STDP learning rule hold little significance. This is because the homeostatic process plays a crucial role in re-scaling synaptic weights to elicit spikes from the output neuron under specific conditions.

The results of the experiments show that a network trained with a single pattern acts as a perfect classifier, which only identifies spatial patterns that are identical to the trained one. The ability of the network to select a specific pattern comes from the structure of the network: while excitatory synapses trigger the output neuron, inhibitory connections perform the selection of the pattern during the testing and validation phase.

When two or more patterns are trained on a single network, the accuracy of the identification task depends only on the Hamming Distance between the code words imprinted in the network.

From the analysis of the classification metrics, it is possible to see that the negative prediction in the case of one or two trained patterns is always $1$, which highlights how the absence of an output spike always correctly identifies that the trained pattern is not present.

On the other hand, positive identification spikes depend on the number of ``don't care'' and ``care little'' synapses or bits imprinted in the network. As the number of ``don't care'' or ``care little'' synapses increases, the number of false positives increases, and this reduces the overall accuracy of the identification task.

In the multiple code word experiments, the classification metrics extracted show that the accuracy is linked to the Hamming Distance between code words. However, when one code word is dropped from identification because its Hamming Distance is too high, the negative prediction and specificity parameters receive lower values, but the overall accuracy improves because the number of false positive identifications drastically reduces. Overall, precision and sensitivity decrease rapidly as the number of false positives increases when the Hamming Distance between code words increases.

This article has presented a method to train a spiking neural network to detect spatial patterns which do not have a temporal component.
These kind of patterns may be found, for example, in the analysis of computer network traffic packets, where each single packet does not have a temporal component, but holds enough information for a neural network to determine the type of traffic that is carries \citep[e.g.:][]{Rasteh2022EncryptedNetwork, Aceto2021DISTILLER:Learning}.
Performing such analysis would require to extract bits of information from the packet (from the header and/or from the payload) and encoding such information following the methodology presented above.
Following the training, this methodology allows the extraction of features ideal for packet classification.
Moreover, using multiple of these classifiers and considering that traffic streams have a typical sequence of packets in a stream, by using a technique such as polychronization \citep{Izhikevich2006Polychronization:Spikes} it would be possible to identify the sequence and determine to which stream it belongs.

An additional example of spatial pattern is represented by static images which can be encoded into spikes to allow a spiking neural network to perform pattern matching tasks on trained patterns \citep[e.g.:][]{Davies2021ANeuro-Robotics}.

In these applications, it is conceivable to adjust the homeostatic factor, calculated during the training phase, to also serve as a similarity factor in matching the input pattern with the trained pattern. Even in the case of a single trained pattern, increasing the homeostatic factor appropriately may enable the network to identify patterns with a certain degree of similarity, rather than requiring an exact match between the input and the trained pattern.

\section*{Acknowledgement}

For the purpose of open access, the authors have applied a Creative Common Attribution (CC BY) licence to any Author Accepted Manuscript version arising from this submission.

This work has been supported by the Department of Computing and Mathematics at the Manchester Metropolitan University and authorised through ethical review 59273.

Development of SpiNNaker software was supported by the EU ICT Flagship Human Brain Project which has received funding from the European Union's FP7 programme under Grant Agreement no. 604102, and from the European Union's Horizon 2020 research and innovation programme under FPA No 650003 (HBP-785907).

Prof. Di Nuovo acknowledges the support of the UK Engineering and Physical Sciences Research Council (grant number EP/X018733/1 for the project ALDENS), and Innovate UK (grant number 10089807 for the Horizon Europe project PRIMI Grant agreement n. 101120727).

The authors extend their gratitude to Prof. Steve Furber and all members, current and past, of the APT group at the University of Manchester for their invaluable support.

\section*{Data and code availability}
The data and the code used to generate this article are available both on the MMU research data storage servers \citep{Davies2024SpiNNakerDataset} using the Digital Object Identifier (DOI): \href{https://doi.org/10.23634/MMU.00634935}{10.23634/MMU.00634935}, and on GitHub at the URL: \url{https://github.com/sergiodavies/SpiNNakerSpatialLearningCodeAndDataset}.

\section*{Conflicts of interest}
The authors declare that they have no known competing financial interests or personal relationships that could have appeared to influence the work reported in this paper.

\section*{Declaration of Generative AI and AI-assisted technologies in the writing process}
During the preparation of this work the authors used ChatGPT in order to improve readability and language. After using this tool, the authors reviewed and edited the content as needed and take full responsibility for the content of the publication.

\printcredits

\bibliographystyle{cas-model2-names}

\bibliography{references}


\bio{a1}
Sergio Davies is a senior lecturer in the Department of Computing and Mathematics at Manchester Metropolitan University. He received his Laurea (equivalent to MSc Eng) in Telecommunication Engineering from the University ``Federico II'' in Napoli, Italy in 2006. Following his Ph.D. in Computer Science at the University of Manchester in 2012, Sergio continued his research on spiking neural networks as a postdoc in the Human Brain Project until 2016. Transitioning to industry, he took leadership roles in strategic advising as a consultant, and then managed various research projects covering computer hardware design, embedded systems architecture, software development and system networking. In 2019, Sergio returned to academia, initially at Sheffield Hallam University before joining Manchester Metropolitan University. His current research focuses on practical application of spiking neural networks in real-world scenarios.
\endbio

\bio{a2}
Andrew Gait is a Research Software Engineer at the University of Manchester within the Research IT department. Following his PhD completed at the University of Leeds in 2007, he has worked in multiple different departments and institutes across the University of Manchester on various software projects and within multiple cross-disciplinary teams. This has included work on a multi-physics software library, the design of software for use in segmenting medical images, and the development and user support of spiking neural network software as part of the Human Brain Project in the APT group in Computer Science.
\endbio

\bio{a3}
Andrew Rowley is a Senior Research Software Engineer at the University of Manchester within the Research IT department. After completing his PhD in Artificial Intelligence at the University of St. Andrews in 2004, he joined the University of Manchester as a Research Software Engineer.  There he worked on various projects related to Access Grid video conferencing before becoming a Senior Research Software Engineer for NaCTeM working on text mining projects. He then joined the Human Brain Project team in Manchester and was working on the SpiNNaker project designing, building and supporting the sPyNNaker software since 2014.
\endbio

\bio{a4}
Alessandro Di Nuovo is Professor of Machine Intelligence at Sheffield Hallam University. He received the Laurea (MSc Eng) and the PhD in Informatics Engineering from the University of Catania, Italy, in 2005 and 2009, respectively. He is  the leader of the Smart Interactive Technologies research laboratory of the Department of Computing. He has published over 120 articles in computational intelligence and its application to cognitive modelling, human-robot interaction, computer-aided assessment of intellectual disabilities, and embedded computer systems. Prof. Di Nuovo has an extensive track record of leading interdisciplinary research and innovation in fundamental and applied topics in AI and Robotics, for which he has received several grants from prestigious funders (EPSRC, European Union) and companies. Currently, Prof. Di Nuovo is editor-in-chief (topics AI in Robotics; Human Robot/Machine Interaction) of the International Journal of Advanced Robotic Systems (SAGE). He is serving as Associate Editor for IEEE Journal of Translational Engineering in Health and Medicine.
\endbio

\end{document}